\documentclass{article}

\PassOptionsToPackage{numbers, compress}{natbib}
    \usepackage[preprint]{neurips_data_2023}
\usepackage[utf8]{inputenc} % allow utf-8 input
\usepackage[T1]{fontenc}    % use 8-bit T1 fonts
\usepackage{hyperref}       % hyperlinks
\usepackage{url}            % simple URL typesetting
\usepackage{booktabs}       % professional-quality tables
\usepackage{amsfonts}       % blackboard math symbols
\usepackage{nicefrac}       % compact symbols for 1/2, etc.
\usepackage{microtype}      % microtypography
\usepackage{xcolor}         % colors
\usepackage{caption}     % multiple captions in a figure

\usepackage{booktabs}
\usepackage{url}
\usepackage{times}
\usepackage{latexsym}
\usepackage{hyperref}

\usepackage{amsmath,amssymb}
\usepackage{graphicx}
\usepackage{tabularx}
\usepackage{multirow}
\usepackage{pifont}
\usepackage{soul}

\usepackage[toc,page]{appendix}
\usepackage{xcolor}
\usepackage{floatrow}
\usepackage{makecell}
\usepackage{xspace}
\usepackage{colortbl}
\usepackage{enumitem}
\usepackage{wrapfig}
\usepackage{subcaption}

\newfloatcommand{capbtabbox}{table}[][\FBwidth]

\usepackage{arydshln}

\usepackage{minibox}
\usepackage{tabularx,ragged2e}
\usepackage{tabularx} % in the preamble
\usepackage{graphicx}
\usepackage{adjustbox}
\usepackage{printlen}
\usepackage{algorithm}
\usepackage[noend]{algpseudocode}
\usepackage{bbm}

\definecolor{aquamarine}{rgb}{0.5, 1.0, 0.83}

\definecolor{HLGreen}{HTML}{C1F7BB}
\definecolor{HLRed}{HTML}{FFC3C3}

\usepackage{listings}

\newcommand{\frameworkname}{\textsc{Collie}}
\newcommand{\datasetname}{\frameworkname{}-v1}
\newcommand{\datasetsize}{2,080}

\title{\frameworkname{}: Systematic Construction of \\Constrained Text Generation Tasks}

\author{%
  Shunyu Yao\footnotemark[1] \quad Howard Chen\footnotemark[1] \quad Austin W. Hanjie\footnotemark[1] \quad Runzhe Yang\footnotemark[1] \quad Karthik Narasimhan\\
  Department of Computer Science, Princeton University\\
  \texttt{\{shunyuy, hc22, hjwang, runzhey, karthikn\}@princeton.edu}
}

\begin{document}

\maketitle

\renewcommand{\thefootnote}{\fnsymbol{footnote}}
\footnotetext[1]{Equal contribution. Project site with code and data: \url{https://collie-benchmark.github.io}. Collie is a herding dog that can help guide domesticated animals like llamas and alpacas.}

\begin{abstract}

Text generation under constraints have seen increasing interests in natural language processing, especially with the rapidly improving capabilities of large language models. However, existing benchmarks for constrained generation usually focus on fixed constraint types (e.g.\,generate a sentence containing certain words) that have proved to be easy for state-of-the-art models like GPT-4. We present \frameworkname{}, a grammar-based framework that allows the specification of rich, compositional constraints with diverse generation levels (word, sentence, paragraph, passage) and modeling challenges (e.g.\,language understanding, logical reasoning, counting, semantic planning). We also develop tools for automatic extraction of task instances given a constraint structure and a raw text corpus. Using \frameworkname{}, we compile the \datasetname{} dataset with \datasetsize{} instances comprising 13 constraint structures. We perform systematic experiments across five state-of-the-art instruction-tuned language models and analyze their performances to reveal shortcomings. \frameworkname{} is designed to be extensible and lightweight, and we hope the community finds it useful to develop more complex constraints and evaluations in the future.

% \footnote{Data and code are available at \url{https://collie-benchmark.github.io}.}.

% \kn{Mention that current NLP benchmarks to eval textgen have been easily aced by GPT-3/4. We need Eval 2.0 type tasks that mix pure language understanding (which was common in eval 1.0 type benchmarks that the commmunity has previously created) with higher level cognitive abilities like logical reasoning, counting, etc. This is one step towards that - language processing/understanding is no longer something we can treat in isolation from other aspects of intelligence.}

\end{abstract}

\section{Introduction}\label{sec:intro}

% - high level motivation for text gen 2.0
% - existing benchmarks are either synthetic or not systematic, scalable
% - describe our grammar framework, how we can auto extract datasets
% - results

Large language models (LLMs) have proven to be extremely capable of generating coherent and fluent text when provided with high-level prompts, performing excellently on popular benchmarks~\cite{wang2018glue} and finding use in production systems such as GPT-4~\cite{gpt4}. Such capabilities have raised the bar for automated text generation, and allow us to explore more nuanced ways of utilizing LMs to produce text. One such line of inquiry has been the paradigm of constrained text generation, whereby the LM can be asked to adhere to a particular topic~\cite{keskar2019ctrl, dathathri2020pplm}, or avoid using certain words~\cite{lu2021neurologic, lu2022neurologicAstar}. However, these works are scratching the surface of a broader phenomenon --- language models are not just about purely generating text anymore, as evidenced by their use in more structured tasks like problem solving~\cite{yao2022react}, code generation~\cite{chen2021codex} and even tool use through API calls~\cite{schick2023tool}. 

This raises a natural question --- `\textit{what should be the next iteration of text generation benchmarks for evaluating large language models that can capture these advanced capabilities}'? We posit that one direction is to natively build in logical and compositional challenges into text generation, through the lens of constrained text generation. Existing benchmarks for constrained generation however focus only on a particular constraint type, require tailored pipeline to collect data and annotations, and/or can only evaluate a specific aspect of the LM's strength~\cite{lin2020commongen, chen2022cognac}. They also suffer from challenges in constructing datasets for scalable and comprehensive evaluation of LLMs.

% Instructing language models (LMs) to follow constraints is a long-standing challenge in text generation. 
% Generating text under constraints is a long-standing and important problem in natural language processing. 
% The constrained generation tasks are common in real-world applications such as assistive writing where users intend to contain specified content or customer support service where business requires to avoid unprofessional responses. More generally, ensuring LMs to follow constraints entails safety implications \cite{}.
% Despite recent progress in scaling and tuning LMs that unleashes their surprising capabilities, having them generate fluent text while satisfying multiple constraints remain difficult \cite{}. 
% Typically, these benchmarks focus on a particular constraint type, requires tailored pipeline to collect data and annotations, and can only evaluate a specific aspect of the LM's strength.

In this paper, we propose \frameworkname{}, a grammar-based framework that enables systematic construction of compositional constraints over diverse generation levels (e.g., words, sentences, paragraphs) and semantic requirements (e.g., language understanding, logical reasoning, counting). Operationally, \frameworkname{} allows researchers to 1) easily specify constraint templates, and then automatically 2) extract constraint values from language corpora, 3) render them into natural language instructions, and 4) evaluate model generations against the constraint instructions.
% \aw{seems to be details that we refer to later on anyways, and second part kind of repeats what was said above.}
% Evaluation done by `parsing' the generated text using the grammar and programmatically evaluating its conformance. 
% In addition to the grammar specification for \frameworkname{}, we develop tools for automatic extraction of test instances given a constraint template and a raw text corpus, reducing the need for repeated constraint data collection, while allowing us to systematically generate diverse constraints, instructions and evaluation scripts at scale. 
The modular and extensible design of \frameworkname{} allows the broader NLP community to contribute additional constraints that can co-evolve with LLM capabilities over time, as well as provide a convenient endpoint for users that only want to evaluate their model without developing their own constraints. %\sy{shorten} \kn{done}

\begin{figure}[t]
    \centering
    \includegraphics[width=0.98\textwidth]{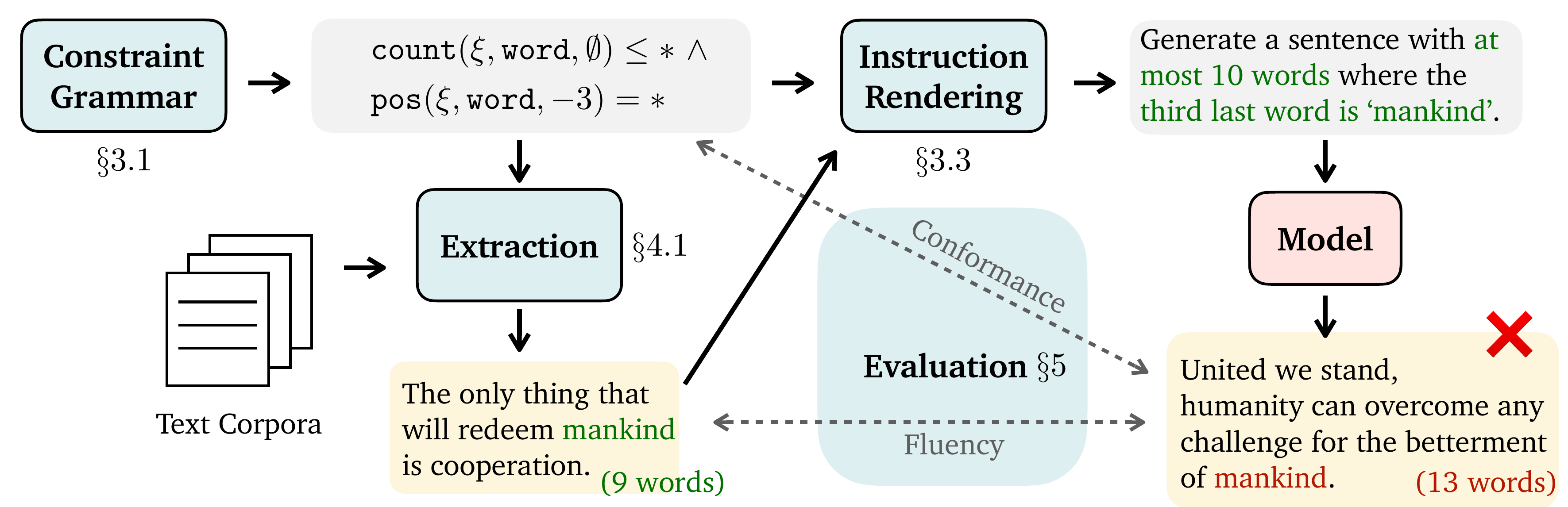}
    \caption{
        Our \textbf{\frameworkname{} framework} for constraint structure specification, ground truth extraction, instruction rendering, and evaluation. First, the user specifies the constraint structure without a specific target value (expressed in $*$). Second, the constraint structure is used to extract ground truth examples from text corpora that contain the target values. Third, the constraint structure and target values are rendered into a natural language instruction. Finally, the model's generation is evaluated against the constraint and the ground truth. The model (\texttt{gpt-3.5-turbo}) violates the constraints by exceeding word limits and leaving the word `mankind' at the end instead of the specified position.
    }
    \label{fig:teaser}
\end{figure}
% \vspace*{-3pt}

% To demonstrate the usefulness of our \frameworkname{} framework, 
We construct the dataset \datasetname{} with \datasetsize{} constraint instances across 13 different types, using three different corpora: Wikipedia~\cite{wikidump}, CC-News~\cite{Hamborg2017ccnews}, and Project Gutenberg~\cite{brooke2015gutentag}. We perform zero-shot evaluations of five state-of-the-art LLMs of varying sizes including
GPT-4~\cite{gpt4} and PaLM~\cite{anil2023palm}.
% GPT-4~\cite{gpt4}, GPT-3.5, PaLM~\cite{anil2023palm}, Vicuna~\cite{vicuna2023}, Alpaca~\cite{alpaca}.
While GPT-4 comparatively performs the best, it still achieves an average constraint satisfaction rate of only {50.9\%}.
% demonstrating the challenging nature of the benchmark.
% We find that all the models struggle to follow counting and positional constraints, and achieve close to 0\% performance on logical compositions.
We find that challenges correlate with position -- for instance, instructing models to begin a sentence with a specific word leads to a 100\% success rate for GPT-4, while asking models to end a sentence with a particular word results in a success rate of 40\%-60\%. 
% Furthermore, requesting the model to use a specific word at an arbitrary position yields an extremely low success rate ($\sim 0\%$). 
% \kn{any other interesting findings to highlight?}

To summarize, we make the following contributions: (1) We introduce \frameworkname{}, a  framework for systematic generation of compositional constraints, that is flexible and extensible. (2) We use \frameworkname{} to curate a new dataset \datasetname{} comprising of 13  constraint structures. (3) We perform a comprehensive evaluation of five state-of-the-art LLMs of varying sizes and provide useful insights for both model and benchmark development in the future.
% \end{enumerate}
%Constrainted text generation: why it's important (align llms to our needs, real-world appications usually have constraints), and what's the problem of existing benchmarks (usually focus on a particular constraint type, specific pipeline for data collection, extraction, evaluation; become too easy; ...)

%\sy{
%We propose \frameworkname. Core idea: unify different constraints using a grammar framework. (briefly introduce the grammar, using table 1) Conceptual benefits:
%\begin{enumerate}
%    \item 0. Grammar naturally allows rich compositions of constraints, stemming from an extrmeley simple core (cnt/position)
%    \item 1. Systematically expresses/categorize/construct a rich set of different constraints
%    \item 2. Scalable pipeline to extract data, construct prompt, evaluate output, etc. with ever-harder constraints (i.e. do not need to write extract/construct/evaluatue code for each new constraint)
%\end{enumerate}

%We did experiments on ... and find ...
%\begin{enumerate}
%    \item 1. compare models: gpt-4 is still much better than ... and those claim as good as chatgpt is not ...
%    \item 2. compare constraints: letter and passage is hard for even gpt-4, etc.....
%\end{enumerate}
%}

\section{Related Work}\label{sec:related}

%\hc{
%\begin{enumerate}
%    \item Controllable text generation (e.g., \cite{keskar2019ctrl, dathathri2020pplm, li2022diffusionlm})
%    \item Constraints in instructions (e.g., \cite{ouyang2022instructgpt}, Cognac)
%    \item Systematic generalization (e.g., \cite{li2022quantifying})
%    \item Planning \& logic (\cite{zhiting2016harness})
%\end{enumerate}
%}
%\aw{I think we should prob discuss DynaBench (benchmarks evolving with models).}
\textbf{Constrained text generation.} 
%Traditional work focuses on decoding strategies to solve simple tasks that control or constrain variables such as ... 
Early work in controllable text generation used control codes to steer the generation towards desired topics or to reduce undesirable content, by controlling for broad attributes such as sentiment or toxicity~\cite{hu2017toward, keskar2019ctrl, dathathri2020pplm, krause2021gedi}. 
%The task typically contains only one or two control attributes and therefore having little emphasis on planning during generation.
Other work on constrained decoding provides to the language model a collection of lexical items as constraints to be included or excluded in the final generated text \cite{hokamp2017lexically, hasler2018neural, dinu2019training, hu2019improved, lin2020commongen, lu2021neurologic, lu2022neurologicAstar, li2022diffusionlm}. 
%\hc{Add many more citations. CommonGen, DiffusionLM, etc}
%While fine-grained control tasks require moderate planning to maintain both fluency and constraint conformance, the constraints are mostly simple such as including words in a sentence.
% Recently, more complex controls such as syntax tree and part-of-speech parses have been explored \cite{li2022diffusionlm}. However, they remain simplistic at a sentence level, leaving the state-of-the-art LLMs unchallenged. \todo{mention work like CommonGen and its related work}
%\hc{I remember seeing some mentions about GPT-4 can do parsing or generate parse conformant text.}
% \paragraph{Instructing with constraints.}
Recent advances in instruction tuning LLMs \cite{ouyang2022instructgpt} have brought major improvements to controllability. These advancements have made it challenging to use existing controllable generation datasets to fully assess the capabilities of modern LLMs.
%This new paradigm provides a unified text interface for expressing constraints without needing any specialized control codes. However, the major focus has been put on taming and instructing the large pretrained model instead of testing their limit when the instruction contains complex constraints.

% \paragraph{Benchmark.}
% Finally, our proposed framework adopts an iterative design close to \citet{kiela2022dyna}. This human-in-the-loop paradigm enables adaptation to the fast growing model capability and can leverage the expressiveness of the constraint grammar to easily expand the dataset.
% \hc{Talk more about the pipeline aspect of the thing.}

\textbf{Grammar-based compositional tests.} 
Building benchmarks with data synthesized from grammars has been explored previously in the context of question answering~\cite{weston2015towards}, instruction following~\cite{chevalier-boisvert2018babyai,ruis2020gscan}) and visual reasoning~\cite{johnson2017clevr}. These benchmarks showcased the utility of grammars to systematically generate a comprehensive set of test cases within each domain. However, these datasets were all synthetic with limited linguistic diversity and practical applicability to real-world scenarios. In contrast, since our \frameworkname{} framework extracts values and examples from natural language corpora to construct the constraints, it represents a more realistic challenge for modern LLMs.

%\cite{brown2020gpt3}

\textbf{Systematic and scalable language benchmarks.}
The emergence of increasingly powerful general-purpose language models has created a need for scalable benchmarks that can systematically and comprehensively evaluate them. A few recent examples include HELM~\cite{liang2022holistic}, BIG-Bench~\cite{srivastava2022beyond}, MMLU~\cite{hendrycks2020measuring}, TaskBench500~\cite{li2022quantifying}, and Natural Instructions~\cite{wang2022super}. However, building such benchmarks require considerable human effort, and may become obsolete when stronger models enter the arena. We provide a new perspective in this race between model capabilities and challenging benchmarks: leverage compositionality to construct automatic and scalable benchmarks with minimal human effort that can co-evolve with model capabilities to remain challenging and relevant.
% In this way, we can enable arbitrarily hard problems while remaining expressive enough to evaluate diverse sets of research challenges such as planning and reasoning.

% Systematic generalization has been studied with the goal to ensure real understanding of tasks by testing the models on compositional tasks \cite{ruis2020benchsys, li2022quantifying}.

% %Benchmarks often constructed challenging examples compositionally. 
% However, the capabilities required to understand compositionally difficult examples do not equate the ability to generate based on compositional constraints that requires planning, which our framework uniquely identifies to be challenging.
% %\citet{valmeekam2022llmcannotplan} proposed to 

% \paragraph{Systematic generalization \& planning.}
% Systematic generalization has been studied with the goal to ensure real understanding of tasks by testing the models on compositional tasks \cite{ruis2020benchsys, li2022quantifying}.
% %Benchmarks often constructed challenging examples compositionally. 
% However, the capabilities required to understand compositionally difficult examples do not equate the ability to generate based on compositional constraints that requires planning, which our framework uniquely identifies to be challenging.
% %\citet{valmeekam2022llmcannotplan} proposed to 
\section{\frameworkname: A Grammar-based Framework for Constrained Text Generation}\label{sec:grammar}
\vspace{-5pt}

% In this section, we describe the grammar for constructing constraints and rendering them as instruction prompts for text generation tasks. The constraints can be either \textit{single-level} or \textit{multi-level}, which are applied to the generated texts to assess whether they meet specific criteria or properties. We formalize the rules for our constraint construction as a context-free grammar (CFG).

\frameworkname{} allows researchers to easily 1) specify textual constraint structures via a grammar, then automatically 2) extract constraint values from text corpora, 3) render constraints into natural language instructions, and 4) evaluate generations with respect to constraints.  
% The use of a grammar provides a systematic process for generating constraints, while also allowing for complex compositions involving several atomic constraints. Each constraint has two forms --  a derivation under the grammar, and a corresponding textual instruction. The textual instruction is used to prompt an LLM into generating text, which can then be `parsed' by the grammar and compared to the derivation to check if it satisfies the constraint.
% \vspace{-10pt}

\renewcommand{\eqref}[1]{(Eq.~\ref{#1})}

\textbf{Grammar.}
Two observations about text constraints motivate a grammar characterization: 1) they involve different \textit{levels} of text, e.g.\,character, word, sentence, or paragraph; and 2) many of them specify either the \textit{count} or \textit{position} at a certain text level (\textit{existence} is equivalent to \textit{count}$~>0$).

Let capitalized letters ($S, M, C, T$) will denote non-terminal variables, and other symbols ($\ell, \circ, \oplus, v$) denote terminals. 
A full \textbf{constraint specification} within our grammar $S$ \eqref{eq:S} consists of two parts: a {generation level} $(\texttt{level}(\xi) = \ell)$ specifying whether the generated text $\xi$ should be a word, a sentence, a paragraph, or a document, and a \textbf{multi-constraint} $M$ \eqref{eq:M}, which is a logical composition of one or more \textbf{base-constraints} $C$. A \textbf{text} $T$ \eqref{eq:T} within these constraints can either be the full generated text $\xi$, or a part of it when qualified with a $\texttt{pos}(\cdot)$. For example, $\texttt{pos}(\texttt{pos}(\xi, \text{paragraph}, 3), \text{sentence}, -1)$ means ``the last sentence of the 3rd paragraph of the generated text''. For the terminal variables, we define a \textbf{level} $\ell$ of a text \eqref{eq:l}, a string or number \textbf{relation} $\circ$ or $\oplus$ \eqref{eq:r}, and a string or number \textbf{value} $v_{\text{str}}$ or $v_{\text{num}}$ \eqref{eq:v} intuitively. $\wedge$ represents the logical `and' operator, and $\vee$ represents the logical `or'. With these definitions, we construct the following grammar:
\begin{align}
\label{eq:S} S & \to (\texttt{level}(\xi) = \ell) \wedge M && \text{(constraint specification)}  \\
\label{eq:M} M & \to C \mid C \wedge M \mid C \vee M &&\text{(multi-constraint)} \\
%\label{eq:C} C & \to \text{Count}(T, \ell, v_{\text{str}}) \circ_{\text{num}} v_{\text{num}} \mid pos(T, \ell, v_{num}) \circ_{str} v_{str}  &&\text{(unit-constraint) } \\
\label{eq:C} C & \to \texttt{count}(T, \ell, v_{\text{str}} \mid \ell'  ) \oplus v_{\text{num}} \mid \texttt{pos}(T, \ell, v_{\text{num}}) \circ v_{\text{str}}  &&\text{(base-constraint) } \\
\label{eq:T} T & \to \xi \mid \texttt{pos}(T, \ell, v_{\text{num}}) && \text{(text)} \\
\label{eq:l} \ell & \to \text{char} \mid \text{word} \mid \text{sentence} \mid \text{paragraph} \mid \text{passage} && \text{(level)} \\
\label{eq:r}  \circ & \to = \mid \ne \quad \quad \oplus  \to = \mid \ne \mid > \mid < \mid \leq \mid \geq && \text{(relation) } \\
\label{eq:v} v_{\text{str}} & \in \Sigma^* \quad  \quad \quad v_{\text{num}} \in \mathbb{Z} && \text{(value) }
\end{align}
At the core of our grammar, we consider two (symmetrical) types of base-constraints $C$ \eqref{eq:C}:
% \begin{enumerate}[]
% \item 

\textbf{1. Count constraints.} $\texttt{count}(T, \ell, v_{\text{str}}) \oplus v_{\text{num}}$ constrains the occurrences of a particular level-$\ell$ string $v_{\text{str}}$. For example, $\texttt{count}(T, \text{word}, \text{`happy'}) \le 3$ means ``T should contain the word `happy' no more than 3 times''. 
In contrast, $\texttt{count}(T, \ell, \ell') \oplus v_{\text{num}}$ constrains the occurrences of level-$\ell$ strings in each level-$\ell'$ unit of text T.
For example, $\texttt{count}(T, \text{char}, \text{sentence}) = 50$ means ``each sentence of text T should have exactly 50 characters''.

\textbf{2. Position constraints.} $\texttt{pos}(T, \ell, v_{\text{num}}) \circ v_{\text{str}}$ specifies that a particular part of the text T should equal (or not equal) the given string $v_{\text{str}}$. For example, $\texttt{pos}(T, \text{word}, 3) = \text{`happy'}$ means ``the 3rd word should be `happy' in text T''. We also allow negative indices for reverse counting, e.g.\,$\texttt{pos}(T, \text{char}, -1) \neq \text{x}$ means ``the last letter should not be `x' in text T''.
% \end{enumerate}

Note that the grammar above can easily be extended to accommodate more types of base-constraints (e.g. part of speech, sentiment) by implementing the corresponding semantic checks --- we leave this to future work. Also for convenience, we use \textbf{constraint structure} to refer to a family of constraint specifications that only differ in their values (e.g.\,\textit{generate a sentence with exactly $x$ words}, $x\in\mathbb{N}$), and \textbf{constraint} to refer to a particular constraint specification with concrete values (e.g.\,\textit{generate a sentence with exactly 5 words}).

% \sy{possibility of extending to more base-constraints (e.g. PoS, ...): not hard, just need corresponding ``semantics'' implemented}

% \section{Compositional Constraints}\label{sec:composition}
\textbf{Examples and conceptual challenges.} %\label{sec:example}
%\hc{Show compositional constraint examples to demonstrate that we can lift the constraints to a higher level (beyond simple controllable generation tasks).}
%\todo{Tony can you help with this section also?}
% \ry{I propose to use this section to go through all example constraints (different generation/target levels, complexity). We can discuss the conceptual difficulties of these example constraints}
% \sy{it flows better to talk about examples here?}
Our grammar can express a wide range of constraints through logical compositions of base-constraints across different text levels. Table~\ref{tab:example_constraints} illustrates some constraint structures across generation levels, identified by names such as \texttt{para01} for paragraph generation, etc.
In addition to the generation levels, \texttt{count} and \texttt{pos} across different levels introduce a variety of challenges. For example, 
\texttt{word01} and \texttt{sent01} challenge token-based language models to count characters; 
\texttt{pass01} requires high-level semantic planning for models to generate a coherent passage under constraints;
\texttt{sent04} and \texttt{para02} challenge models to generate text with presence or absence of particular words; 
\texttt{sent03}, \texttt{para03}, and \texttt{para04} require counting at multiple levels; 
and \texttt{word02}, \texttt{word03}, \texttt{sent02}, \texttt{para05}, and \texttt{pass01} combine counting and positional challenges at different levels, which can be considered most demanding conceptually.
% These tasks are considered the most conceptually demanding. 
We empirically assess the difficulty of  constraint structures in Section~\ref{sec:results}.

\begin{table}[thbp]
    \footnotesize
    \centering
        \resizebox{\columnwidth}{!}{%

    \begin{tabular}{
        >{\raggedright\arraybackslash}p{0.6cm}
        >{\raggedright\arraybackslash}p{5.5cm}
        >{\raggedright\arraybackslash}p{6.9cm}
        % >{\raggedright\arraybackslash}p{1.8cm}
        % >{\raggedright\arraybackslash}p{1.2cm}
        % >{\raggedright\arraybackslash}p{1cm}
    }
        \hline
        \textbf{ID} & \textbf{Example instruction} & \textbf{Multi-constraint $M$} 
        % & \textbf{Composition complexity} 
        % & \textbf{Difficulty}
        \\
        \hline
        
        word01 & 
        Generate a word with at least 15 letters. & 
        $\texttt{count}(\xi, \text{char}, \text{word}) \ge 15$ 
        % & atomic 
        % & counting
        \\
        
        \hline
        word02 & 
        Generate a word with 10 letters, where letter 1 is `s', letter 3 is `r', letter 9 is `e'.
        &  $\texttt{count}(\xi, \text{char}, \text{word})=10$ 
        $\wedge$ $\texttt{pos}(\xi, \text{char}, 1)=\text{`s'}$ 
        $ \wedge$ $ \texttt{pos}(\xi, \text{char}, 3)=\text{`r'}$
        $\wedge$  $\texttt{pos}(\xi, \text{char}, 9)=\text{`e'}$
        % &  logical, single-level 
        % & counting, position
        \\
        
        \hline        
        word03 & 
        Generate a word with at most 10 letters and ends with ``r". & 
        $\texttt{count}(\xi, \text{char}, \text{word})\leq 10$  $\wedge$ $ \texttt{pos}(\xi, \text{char}, -1)= \text{`r'}$
        % & logical, single-level 
        % & counting, position
        \\

        \hline      
        sent01 & 
        Please generate a sentence with exactly 82 characters. Include whitespace into your character count. & 
        $\texttt{count}(\xi, \text{char}, \text{sentence}) = 82$ 
        % & 
        % & 
        \\

        \hline      
        sent02 & 
        Generate a sentence with 10 words, where word 3 is “soft” and word 7 is “beach” and word 10 is “math”. & 
        $\texttt{count}(\xi, \text{word}, \text{sentence}) = 10 $ 
        $\wedge$ $\texttt{pos}(\xi, \text{word}, 3)=\text{``soft"}$ $\wedge$ $\texttt{pos}(\xi, \text{word}, 7)=\text{``beach"}$ $\wedge$ $\texttt{pos}(\xi, \text{word}, 10)=\text{``math"}$
        % & 
        % & 
        \\
        
        \hline
        sent03 & 
        Generate a sentence with at least 20 words, and each word less than six characters. & 
        $\texttt{count}(\xi, \text{word}, \text{sentence}) \ge 20 $ $\wedge $ $\texttt{count}(\xi, \text{char}, \text{word}) \le 6$
        % $\texttt{count}(\xi, \texttt{word}, \emptyset)\geq 20 \wedge \texttt{count}(\xi,\texttt{char}, \texttt{pos}(\xi, $
        % $\texttt{word}, 0)) < 6 \wedge \texttt{count}(\xi,\texttt{char}, \texttt{pos}(\xi,$
        % $\texttt{word}, 1)) < 6 \wedge ...$ 
        % & 
        % & 
        \\

        \hline
        sent04 & 
        Generate a sentence but be sure to include the words “soft”, “beach” and “math”.  & 
        $\texttt{count}(\xi, \text{word}, \text{`soft'})> 0$ $\wedge$ $\texttt{count}(\xi, \text{word}, \text{`beach'})> 0$ $\wedge$ $\texttt{count}(\xi, \text{word}, \text{`math'}) > 0$ 
        % & 
        % & 
        \\

        \hline
        para01 & 
        Generate a paragraph where each sentence begins with the word “soft”.  & 
        $\texttt{pos}(\texttt{pos}(\xi, \text{sentence}, 1), \text{word}, 1) = \text{`soft'} \wedge$
        $\texttt{pos}(\texttt{pos}(\xi, \text{sentence}, 2), \text{word}, 1) = \text{`soft'}
        \wedge ...
        $
        % & 
        % & 
        \\
        \hline
        para02 & 
        Generate a paragraph with at least 4 sentences, but do not use the words “the”, “and” or “of”.  & 
        $\texttt{count}(\xi, \text{sentence}, \text{paragraph}) \geq 4$ $\wedge$ $\texttt{count}(\xi, \text{word}, \text{`the'})=0$ $\wedge$ $\texttt{count}(\xi, \text{word}, \text{`and'})=0$ $\wedge$ $\texttt{count}(\xi, \text{word}, \text{`of'})=0$
        % & 
        % & 
        \\
        \hline
        para03 & 
        Generate a paragraph with exactly 4 sentences, each with between 10 and 15 words. & 
        $\texttt{count}(\xi, \text{sentence}, \text{paragraph}) = 4$ $\wedge$ $\texttt{count}(\xi, \text{word}, \text{sentence}) \ge 10$ $\wedge$ $\texttt{count}(\xi, \text{word}, \text{sentence}) \le 15$
        % & 
        % & 
        \\
        \hline
        para04 & 
        Generate a paragraph with at least 3 sentences, each with at least 15 words. & 
        $\texttt{count}(\xi, \text{sentence}, \text{paragraph}) \ge 3 $ $\wedge $ $\texttt{count}(\xi, \text{word}, \text{sentence}) \ge 15$
        % & 
        % & 
        \\
        \hline      
        para05 & 
        Generate a paragraph with 2 sentences that end in “math” and “rock” respectively. & 
        $\texttt{count}(\xi, \text{sentence}, \text{paragraph}) = 2 $ $\wedge $ 
        $\texttt{pos}( \texttt{pos}(\xi, \text{sentence}, 1), \text{word}, -1)=\text{``math"} $ $\wedge $
        $\texttt{pos}( \texttt{pos}(\xi, \text{sentence}, 2), \text{word}, -1)=\text{``rock"}$
        % & 
        % & 
        \\

        \hline      
        pass01 & 
        Generate a passage with 2 paragraphs, each ending in “I sit.” and “I cry.” respectively. & 
        $\texttt{count}(\xi, \text{paragraph}, \text{passage}) = 2 $ $\wedge $ 
        $\texttt{pos}( \texttt{pos}(\xi, \text{paragraph}, 1), \text{sentence}, -1)=\text{``I sit."} $ $\wedge $
        $\texttt{pos}( \texttt{pos}(\xi, \text{paragraph}, 2), \text{sentence}, -1)=\text{``I cry."}$
        % & 
        % & 
        \\
        \hline
    \end{tabular}
    }
    \caption{List of all constraint structures used in \frameworkname-v1, with (simplified) example values.
    % , with example values (e.g. 82, “soft”). Additional constraints can be easily added using our constraint grammar.
    }
    \label{tab:example_constraints}
    
\end{table}

% \sy{talk about which part of grammar can be further extended? or talk about it later in discussion}

% \subsection{Automatic dataset construction with \frameworkname{}}
In conjunction with the grammar, we develop a set of \textit{compiling tools} to help construct datasets with minimal human efforts.  Concretely, the pipeline of dataset construction involves 4 stages (Figure~\ref{fig:teaser}): 

\textbf{1. Specify constraint structures.} Researchers can specify constraint structures (e.g.\,Table~\ref{tab:example_constraints}), and optionally with a value range (e.g.\,``generate a sentence with $x$ words'', and $5 \le x \le 10$). This is the only stage that involves manual effort.

\textbf{2. Extract constraint values from corpora.} We design an automatic extraction algorithm that runs through a given text corpus to find strings that fit a constraint structure with some value ranges. 
For example, given the constraint structure $\texttt{count}(\xi, \text{word}, \emptyset)= x$ with value range $5 \le x \le 10$, the extraction algorithm returns sentences in the corpus that have 5-10 words, with  associated word counts. 
This ensures each constraint has at least one natural solution. More details are in Section~\ref{sec:constraint_creation}.

\textbf{3. Render natural language instructions.} Each constraint can be rendered into a natural language instruction (Table~\ref{tab:example_constraints}) via ruled-based translation to prompt language models (details in Section~\ref{appendix:render}). 
% combining predefined language fragments while traversing from the root node of the tree. 
% Additionally, we offer the option to enhance the template-based prompts using an LLM for further refinement. 
% \kn{this should clearly explain the process in 1-2 sentences. talk about LLM use if we did.} 

\textbf{4. Evaluate generations.} Given text $\xi$ generated by a model, we use a parser to evaluate it against a constraint specification $S$ and derive a True/False value, indicating if $\xi$ satisfies $S$. We use an average \textit{Success Rate} as the main metric to evaluate constraint conformance. We can also compare the fluency of $\xi$ against the corpus-extracted ``groundtruth'' text, and render more fine-grained natural language feedback indicating which base-constraints are met and which not (see Section~\ref{appendix:feedback}).

\section{Dataset Construction: \frameworkname{}-v1}\label{sec:setup}
% We describe detailed setup and usage of \frameworkname in the following subsections. For convenience, let us define the \textit{constraint structure} to be the family of constraints consisting of all possible target values. (e.g. \textit{Sentence with $x$ words}, where $x\in\mathbb{N}$). Therefore, a constraint $C$ as defined earlier, consists of the constraint structure, together with specific target value(s) (e.g. $x=5$). 

% The intended use of \frameworkname{} consists of two main stages: (1) constraint creation/extraction and (2) model evaluation. The constraints generated from the first two stages can be cached, versioned and released for broader use by the community. The advantages of this two stage approach are threefold: First, the broader NLP community can contribute to future dataset releases by adding additional constraints, leading to metrics that better reflect the interests of diverse stakeholders. Second, the curated constraint set can co-evolve with models to become more challenging and comprehensive as model capabilities improve. Finally, the released constraints remain a convenient endpoint for users that only want to evaluate their model without developing their own constraints. \sy{unify and shorten}

We construct \frameworkname{}-v1 using constraints structures from Table~\ref{tab:example_constraints}, which contains \datasetsize{} constraint instances from 13 constraint types, with 1,435 unique constraint prompts.
The broader NLP community can contribute to future dataset releases by adding additional constraints, metrics, data sources. The curated constraint set can co-evolve with models to become more challenging and comprehensive as model capabilities improve. All code and data are available at \url{https://collie-benchmark.github.io}, and further details on dataset construction are also provided in Section~\ref{appendix:datasets}.
% Finally, the released constraints remain a convenient endpoint for users that only want to evaluate their model without developing their own constraints.

\subsection{Constraint specification and extraction}\label{sec:constraint_creation}

\textbf{Constraint specification.} We begin by defining 13 constraint structures. We chose these 13 structures to span various generation levels (word, sentence, paragraph and passage generation) and challenges (counting, position). In total, we have 3 word-level, 4 sentence-level, 5 paragraph-level, and 1 passage-level constraint structures. Of these 13 constraint structures, 5 are single-level and the remaining 8 are multi-level constraints. See Table \ref{tab:example_constraints} for the exact constraint structures we use.

\textbf{Constraint extraction.} While constructing constraint structures is straightforward using our grammar, choosing constraint targets is challenging for two reasons: (1) Not all targets will admit a conforming natural language string. For instance, the constraint, ``Generate a two word sentence beginning with the word \textit{The}.'' has no grammatically acceptable answer. (2) Even if a constraint admits a \textit{possible} answer, it may not admit a \textit{plausible} answer. For instance, ``Generate a sentence with 1928 words'' is possible, but any such sentence is very unlikely to appear in regular discourse. 

To address both challenges, we sample constraint target values from natural language corpora, which we denote as the \textit{data source}. Given a constraint structure $\mathcal{C}$ and documents $\mathcal{D} = \{d_1,...,d_n\}$, we chunk each document into a series of strings $d_i=\{s_1,...,s_m\}$, where each $s_i$ can be a sentence, paragraph, or passage as required by $\mathcal{C}$. 
Each string $s_i$ undergoes source-specific automated filtering and post-processing to remove artifacts, which we detail in Section~\ref{sec:data_sources}. In all the data sources we consider, a single or double newline is used to delimit paragraphs. To obtain sentences, we use the {\tt nltk} \cite{bird2006nltk} sentence tokenizer on each paragraph. To obtain passages, we string together multiple consecutive paragraphs that survive filtering. To obtain word-level constraint targets, we iterate each $s_i$ from an English language word list\footnote{We use the word list from \url{http://www.gwicks.net/textlists/english3.zip}.}.
Given $\mathcal{C}$ and $s_i$, we extract target values such that $\mathcal{C}$ is satisfied. In most cases, the satisfying target values can be directly extracted using our provided utilities. For example, for constraints with structure ``sentence with $x$ words'', we can directly apply word tokenization and counting to the example string $s_i$. In cases in which direct extraction is not possible, (e.g. ``do not include word $w$''), we specify a range of possible targets (e.g. $\{$\textit{the, and, of}$\}$) to sweep over. 
All in all, our approach ensures that (1) there exists a natural language string that can satisfy each constraint and target pair, and (2) the targets follow a plausible distribution induced by natural language corpora. Our extraction system is extensible, and can operate on new constraints and data sources with minimal modifications. An end user wishing to extract targets from their constraints on our data sources simply provides the constraint configuration and optionally a target range to sweep over, and the extraction is handled automatically using our tools. Adding additional data sources to the extraction pipeline is similarly easy, requiring a text delimiter, and optional string filtering and post-processing functions.

\subsection{Data sources}\label{sec:data_sources}
\begin{figure}[t]
  \centering
  % \begin{subfigure}[b]{0.32\textwidth}
  %   \includegraphics[width=\linewidth]{figs/grouped-dataset-size.pdf}
  %   % \caption{Number of constraints from each constraint structure.}
  %   % \label{fig:dataset_size}
  % \end{subfigure}
  % \hfill
  % \begin{subfigure}[b]{0.33\textwidth}
  %   \includegraphics[width=\linewidth]{figs/frac_filtered.pdf}
  %   % \caption{Fraction of strings removed by automated filtering.}
  %   % \label{fig:frac_filtered}
  % \end{subfigure}
  % \hfill
  % \begin{subfigure}[b]{0.33\textwidth}
  %   \includegraphics[width=\linewidth]{figs/length_stats.pdf}
  %   % \caption{Length statistics for different levels for each data source.}
  %   % \label{fig:length_stats}
  % \end{subfigure}
  \includegraphics[width=\linewidth]{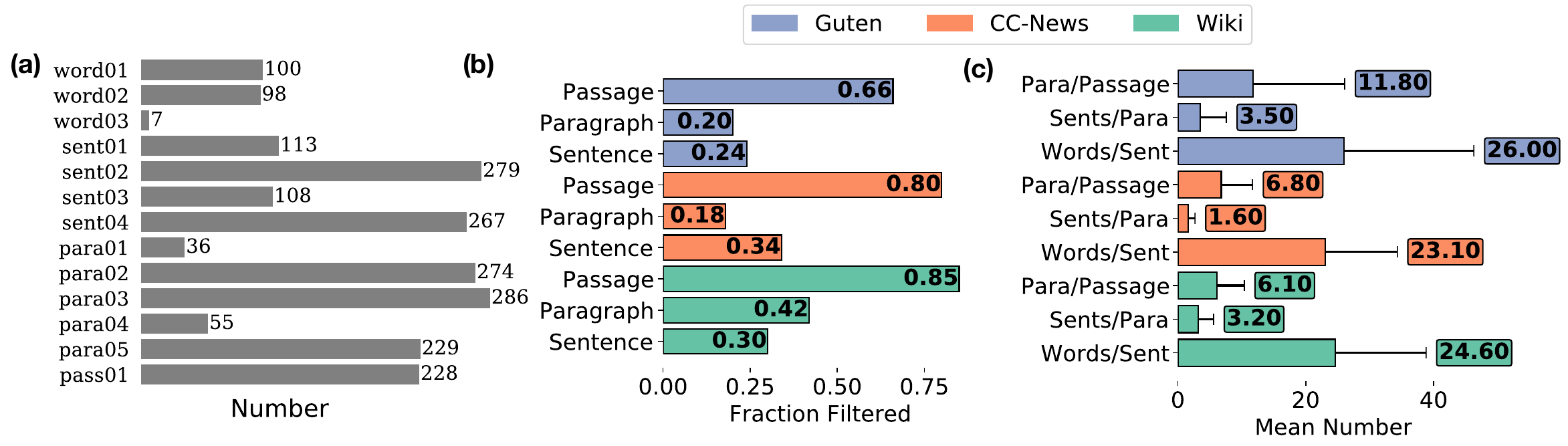}
  \caption{\textbf{Data statistics.} \textbf{(a)} Number of constraints from each constraint structure. \textbf{(b)} Fraction of strings removed by automated filtering. \textbf{(c)} Length statistics for different levels for each data source.}
  \label{fig:vombined_data}
  % \vspace{-1em}
\end{figure}

To ensure adequate coverage of diverse styles and content, we extract constraint targets from three distinct data sources: \textit{Wikipedia} (Wiki)\cite{wikidump}, \textit{Common Crawl News} (CC-News)\cite{Hamborg2017ccnews}, and the \textit{Project Gutenberg Corpus} (Guten)\cite{brooke2015gutentag}. We provide an overview of these data sources below and in Section~\ref{appendix:sources}.

\textbf{Wiki.} Wikipedia (Wiki)~\cite{wikidump} is a  dump from the \textit{Wikimedia Foundation}, consisting of over 6 million English Wikipedia articles. We included this data source for its diverse subject matter. We filter any string that is a figure caption, contains a URL, or does not contain any sentences. Each filter is based off   heuristics and utilize regular expressions (see Section~\ref{appendix:filter}). The strings are then stripped of section titles, which are always followed by a single newline, instead of a double newline.\footnote{We use the preprocessed {\tt 20220301.en} split from Huggingface.}

\textbf{CC-News.} The Common Crawl News corpus (CC-News) \cite{Hamborg2017ccnews} consists of 708,241 English language news articles published between Jan 2017 and December 2019. We include CC-News to include documents which feature interview dialogues, as well as popular culture and current events. In addition to the filters used for Wiki, we found it necessary to include a copyright filter that rejects any string containing the copyright symbol ©, or begins with the word ``copyright''. We do not use any post-processing function for strings from CC-News.

\textbf{Guten.} The \textit{Project Gutenberg} corpus (Guten) \cite{brooke2015gutentag} consists of over 50,000 documents that include fiction, histories, biographies, and other works that are in the public domain in the United States\footnote{We use the processed corpus from \textit{Gutenberg, dammit} at: \url{https://github.com/aparrish/gutenberg-dammit}}. We include this corpus for its variety in style with text from different time periods. Similar to Wiki, we remove any string without a sentence using the same heuristics. We further filter any section titles by removing any line entirely in upper case. Each document in Guten is formatted using Markdown. For each string we remove Markdown formatting (e.g. **, \_\_,), and remove all references and footnotes using regular expressions (see Section~\ref{appendix:postprocess}). We further clean the text by replacing single newlines and multiple consecutive whitespace tokens with a single whitespace token.

\subsection{Data validation and statistics}
We extract constraints from 300 randomly sampled documents from each source. After extracting the target values, we sample up to 100 targets for each constraint structure on each data source. We remove any string targets by that begins or ends with any character that is not a letter or number. 
We randomly sample 5 out of these 100 targets and their supporting examples to qualitatively verify their validity. Since the extraction process is relatively fast, we modify filters and post-processors if there are systemic issues and re-run the extraction phase. We provide statistics of the final number of constraints from each constraint structure in Figure \ref{fig:vombined_data}(a). Some constraints (e.g. number of sentences per paragraph) are tightly clustered around the mean, and thus does not induce many valid constraint targets. 
The fraction of strings filtered for each data source and level is presented in Figure \ref{fig:vombined_data}(b). The automated filtering removes a large fraction of the strings in most cases, as high recall is important to ensure the quality of extracted targets. The high fraction of omitted passages is due to the removal of passages less than two paragraphs in length. Mean length statistics for each level and data source is presented in Figure \ref{fig:vombined_data}(c). There is high variance in lengths across data sources, demonstrating the diversity of the text used to extract the targets. For instance, the mean number of paragraphs per passage is 11.8 in Guten, while it is almost half that in Wikipedia at 6.1 paragraphs per passage. Finally, for \texttt{para02}, \texttt{para03}, and \texttt{pass01} constraints, authors did manual filtering to make sure paragraph and passage examples are clean and with proper lengths.
%Paragraph length is also highly varied among data sources, with CC-News containing 1.6 sentences per paragraph, and Guten containing more than twice as many at 3.5 sentences per paragraph.
\section{Results \& Analysis}\label{sec:results}

\begin{figure}
    \centering
    \includegraphics[width=\textwidth]{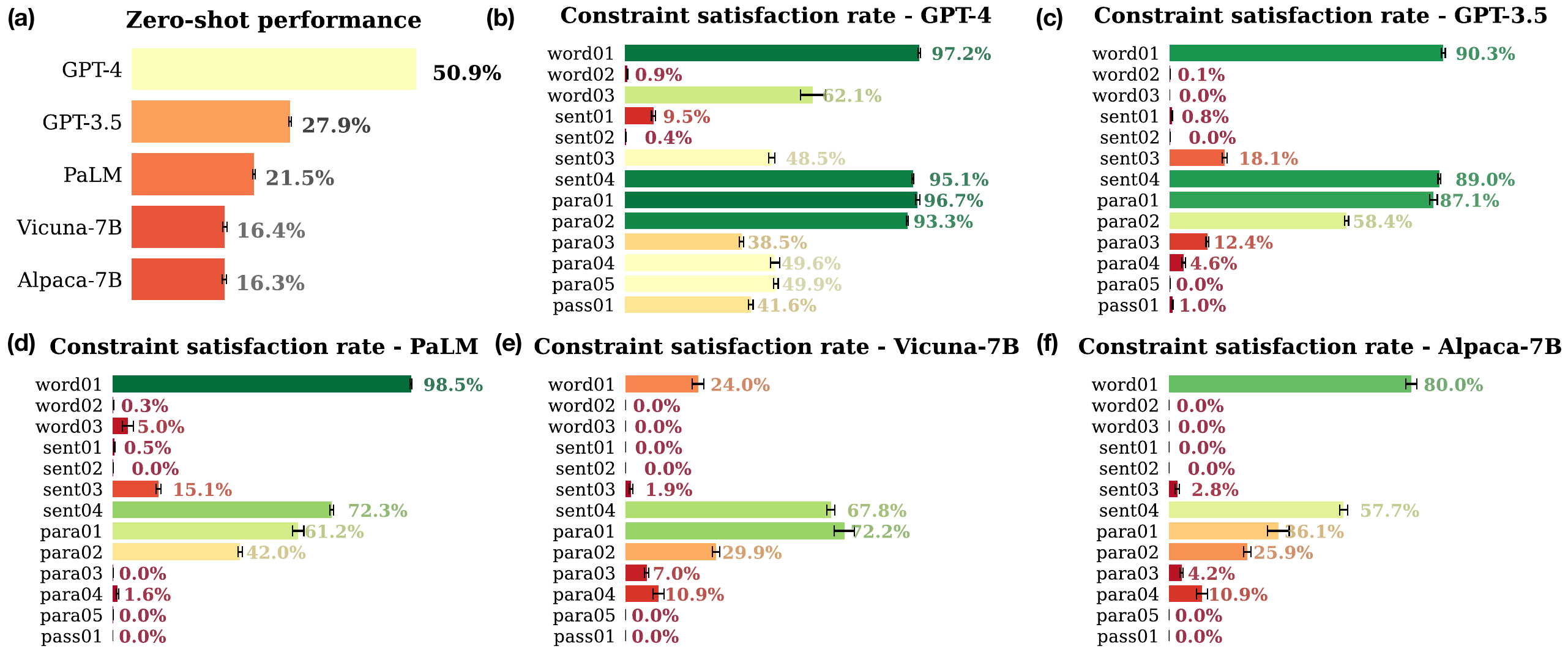}
    \caption{\textbf{Model comparison.} \textbf{(a)} Overall model performance summarized by weighted average across all constraint groups. \textbf{(b)} -\textbf{(f)} Constraint satisfaction rates of generated texts by GPT-4, GPT-3.5, PaLM, Vicuna-7B, and Alpaca-7B across various constraint groups. Error bars represent standard error. Constraint group names are in Table \ref{tab:example_constraints}. Sample sizes are reported in Figure~\ref{sup-fig:grouped-sample-size}.}
    \label{fig:grouped-comparison}
    % \vspace{-0.5em}
\end{figure}

\begin{figure}
    \begin{floatrow}
        \ffigbox[\FBwidth]
            {\caption{\textbf{Pass@k performance.} We sample the model-generated text 20 times for all instruction prompts in the dataset. The curves represent the average pass rate across all instruction prompts up to $k$ samples. The shaded areas indicate the standard errors.}\label{fig:pass-k}}
            {\includegraphics[width=0.45\textwidth]{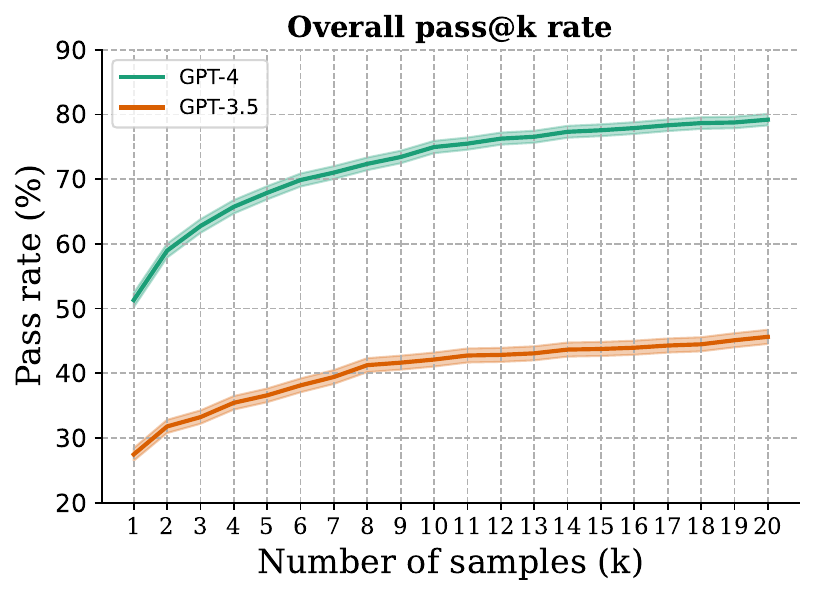}}
        \ffigbox[\FBwidth]
            {\caption{\textbf{Position effect.} Satisfaction rates of LMs on tasks involving $\texttt{pos}(\xi,\text{level}, i)$. The tasks \texttt{word02} and \texttt{sent02} impose constraints on characters and words at arbitrary positions. The task \texttt{para01} constrains the first word. The tasks \texttt{word03}, \texttt{para05}, and \texttt{pass01} constrain the last characters, words, and sentences.}\label{fig:position-effect}}
            {\includegraphics[width=0.55\textwidth]{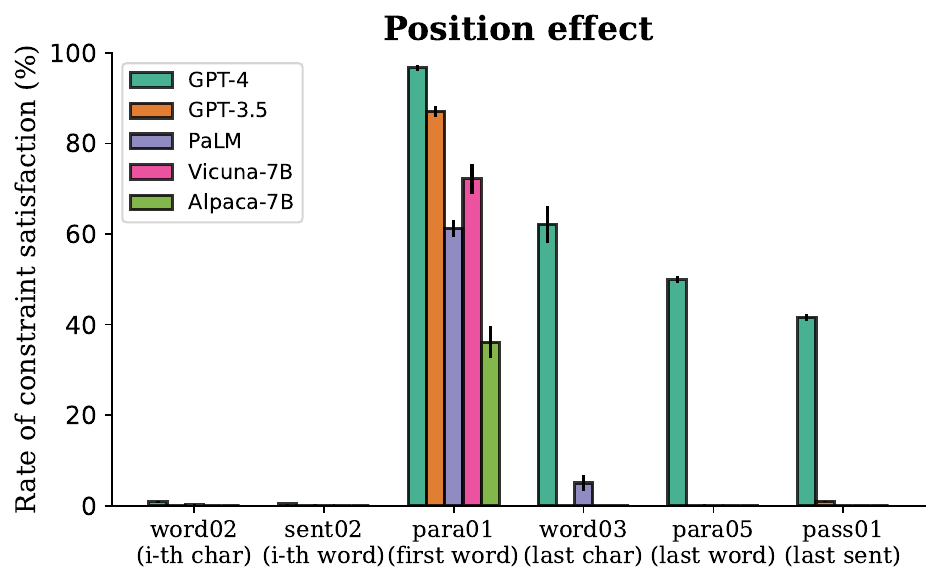}}
    \end{floatrow}
    % \vspace{-1em}
\end{figure}

% \subsection{Experiment setup}

Our main experiments in this paper focus on a zero-shot\footnote{We tried one-shot prompting and find GPT performances similar to zero-shot, see Section~\ref{appendix:oneshot}.} prompting setup with the following language models (LMs): 1) larger and closed-source LMs such as OpenAI GPT~\cite{brown2020gpt3,openai2023gpt4} (\texttt{gpt-3.5-turbo}, \texttt{gpt-4}) and Google PaLM-2~\cite{anil2023palm} (\texttt{text-bison-001}); 2) smaller and open-source LMs such as Alpaca-7B~\cite{alpaca}, Vicuna-7B~\cite{vicuna2023}
% , ChatGLM-6B-1.1~\cite{}, Falcon-7B~\cite{}
. 
By default, we use a sampling temperature of 0.7, and sample multiple trials (20 for GPT/PaLM, 5 for Alpaca/Vicuna). 
% \kn{how many exactly?} 
%
All experiments were run in July, 2023.

\subsection{Main results}

\textbf{Zero-shot performance comparison.} 
% Our results highlight the superior zero-shot performance of GPT-4 to generate text with constraints over other language models. 
As evidenced in Figures \ref{fig:grouped-comparison}(a), GPT-4 consistently surpassed other models in zero-shot constrained text generation performances, achieving more than twice the constraint satisfaction rate than other non-GPT models. The overarching performance trend observed shows GPT-4 leading the pack, followed by GPT-3.5 and PaLM with a large gap, and then followed closely by the smaller models, Vicuna-7B and Alpaca-7B. 
% This trend is illustrated in Figure \ref{fig:grouped-comparison} (a).

\textbf{Constraints all models can follow.} Certain tasks, specifically \texttt{word01} (generating a word with at least $a$ letters), \texttt{sent04} (generating a sentence containing words X, Y, Z), and \texttt{para01} (generating a paragraph with each sentence starting with the word X), posed minimal challenge to the majority of contemporary language models. These tasks demonstrate the proficiency of current models at simple constraints ensuring existence, as depicted in Figure~\ref{sup-fig:performance-comparison}(f).

\textbf{Constraints partially solved by GPT-4 only.} However, a notable distinction arose when tasks incorporated more counting/position constraints and requested longer generations. Tasks such as \texttt{word03}, \texttt{para04}, \texttt{para05}, and \texttt{pass01} were only partially addressed by GPT-4, with constraint satisfaction rates ranging between 40\% and 70\%. Despite GPT-4's partial success in these tasks, other models failed to deliver any satisfactory performance.

\textbf{Constraints remaining very challenging.} Furthermore, some tasks proved challenging across all models. Tasks \texttt{word02}, \texttt{sent01}, \texttt{sent02}, and \texttt{para03} present challenges in terms of arbitrary position constraints and mixed counting levels (see Section \ref{sec:analysis} for detailed analysis), 
% \kn{refresh reader memory on what these tasks are}
indicating areas that necessitate further advancements in language model technology. Moreover, the average pass@20 rate of GPT-4 was above 63\% across all constraints, significantly higher than the 32\% achieved by GPT-3.5, as depicted in Figure \ref{fig:pass-k}. Although GPT-4 demonstrated a significant performance advantage, its constraint satisfaction rate of 63\% is far from perfect. This suggests considerable scope for improvement in controllable text generation with language models. These findings underscore the opportunities and challenges in the continued evolution of language models.

% \ry{will elaborate on the following points}
% \begin{itemize}
%     \item GPT-4 significantly outperforms other models. Fig. \ref{fig:grouped-comparison}, Fig. \ref{sup-fig:performance-comparison}
%     \item Overall trend: GPT-4 > GPT-3.5 $\gtrsim$ PaLM > Vicuna-7B $\approx$ Alpaca-7B. Fig. \ref{fig:grouped-comparison} (a)
%     \item Tasks \texttt{word01} (generate a word with at least $a$ letters), \texttt{sent04} (generate a sentence containing words X, Y, Z), \texttt{para01} (generate a paragraph with each sentence starting with the word X) are easy for most modern language models. Fig. \ref{sup-fig:performance-comparison} (f)
%     \item Only GPT-4 achieves partial success in tasks \texttt{word03}, \texttt{para04}, \texttt{para05}, and \texttt{pass01}, with constraint satisfaction rates ranging from 0.4 to 0.7. However, the other models fail these tasks entirely. 
%     \item Tasks \texttt{word02}, \texttt{sent01}, \texttt{sent02}, \texttt{para03} are difficult for all models.
%     \item The average pass@20 rate of GPT-4 is above 0.63 over all the constraints, while it's less than 0.32 for GPT-3.5. Fig. \ref{fig:pass-k}. Even with a constraint satisfaction rate of 0.63, which is not considered satisfactory, there is clear room for improvement in controllable text generation with language models.
% \end{itemize}

\subsection{Analysis} \label{sec:analysis}

\textbf{Performance consistency across data sources.} We observe a high degree of consistency in the performance of models on a given constraint structure, regardless of the data source. This uniformity is evident across all models, as highlighted in  Figure~\ref{sup-fig:performance-comparison} (g). This indicates that the ability of a language model to adhere to the logic of constraints takes precedence over the specific target values or the distribution of the data.

\textbf{Position effect.} As depicted in Figure \ref{fig:position-effect}, the $\texttt{pos}(\xi, \text{level}, i)$ function, constraining the $i$-th sub-string (letter, word, or sentence), exhibits varying levels of difficulty depending on the value of $i$. Models generally perform well when the positional constraint is applied to the first sub-string ($i=1$, task \texttt{para01}). However, only GPT-4 displays partial success with the last positional constraints ($i=-1$, tasks \texttt{word03}, \texttt{para05}, \texttt{pass01}). Notably, all models encounter difficulties when generating text that satisfies positional constraints at arbitrary positions $i$. Additionally, we find that the position effect exhibits a lower sensitivity to constraint levels.

\begin{figure}
    \centering
    \includegraphics[width=\textwidth]{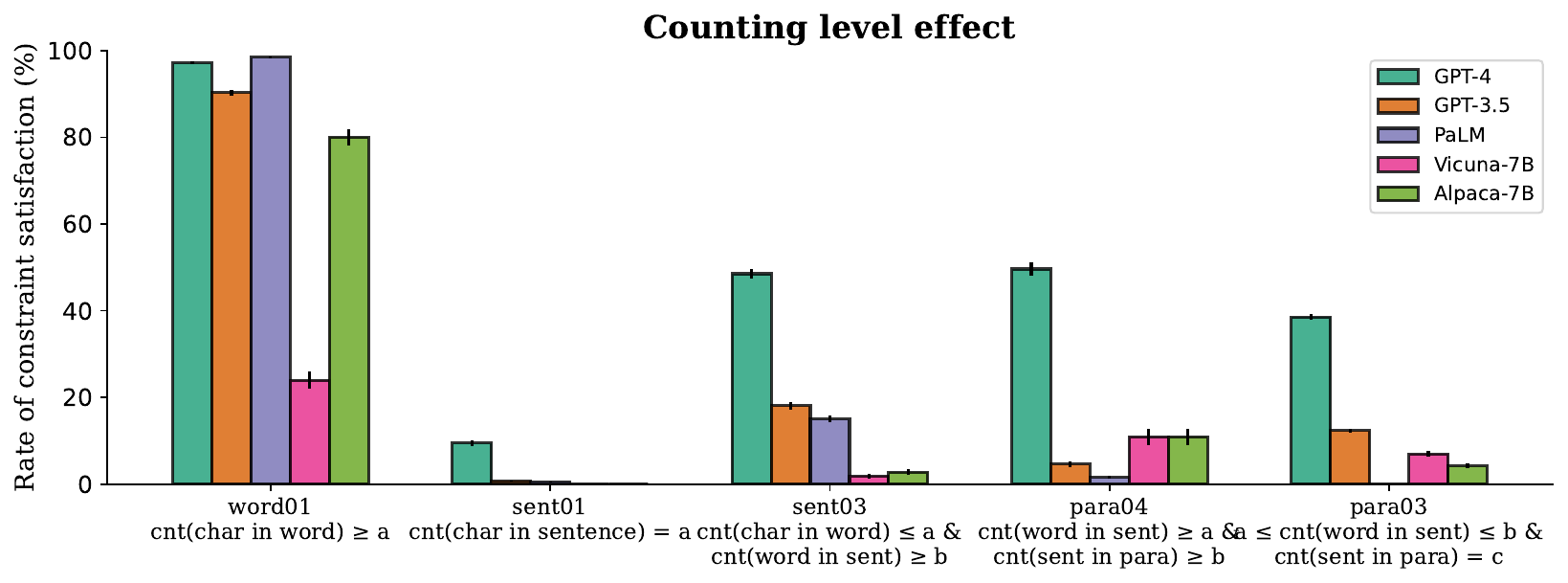}
    \caption{\textbf{Counting level effect.} Satisfaction rates for LMs on tasks involving $\texttt{count}(\xi, \text{level}, \varphi)$. Task \texttt{word01} sets a minimum word length of $a$. Task \texttt{sent01} requires exactly $a$ characters in a sentence. Task \texttt{sent03} asks a sentence to contain at least $b$ words, with each word no longer than $a$ letters. Task \texttt{para04} asks a paragraph to consist of at least $b$ sentences, each containing a minimum of $a$ words. Task \texttt{para03} further imposes an upper limit on the number of words per sentence.}
    \vspace{-1em}
    \label{fig:counting-level-effect}
\end{figure}

\textbf{Counting level effect.} The task of counting characters within a word is comparatively easier for models than counting characters within a sentence, as illustrated in Figure \ref{fig:counting-level-effect}. Furthermore, tasks demanding exact equality (task \texttt{sent01}) prove more challenging than those requiring a range (task \texttt{para03}), and are considerably more difficult than tasks specifying just an upper or lower bound (tasks \texttt{word01}, \texttt{sent03}, \texttt{para04}).

\textbf{Increased difficulty with logical composition.} The incorporation of logical compositions into constraints considerably increases their difficulty. Task \texttt{sent03} serves as an example of this, adding an extra constraint at the sentence level compared to task \texttt{word01}. Despite the assumption that the added constraint should be manageable for all models, performance on task \texttt{sent03} uniformly trails behind that on task \texttt{word01}, as shown in Figure \ref{fig:counting-level-effect}. This highlights the intricacy and challenge introduced by logical compositions within constraints.

\begin{figure}
    \centering
    \includegraphics[width=\textwidth]{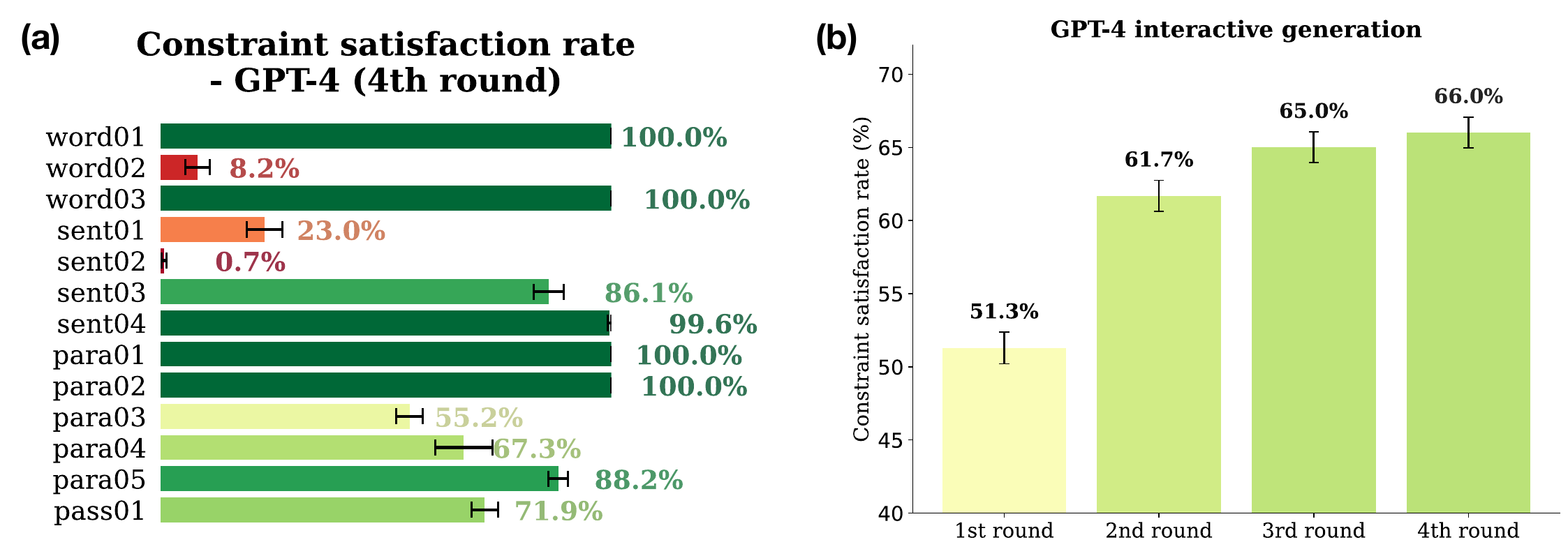}
    \caption{\textbf{GPT-4 interactive generation performance.} \textbf{(a)} Constraint satisfaction rate of GPT-4 generated texts in the 4th round across various constraint groups. \textbf{(b)} GPT-4 overall performance in different feedback rounds. The 1st round is zero-shot, and the 2nd - 4th rounds are with feedback.}
    \vspace{-1em}
    \label{fig:interactive}
\end{figure}

\textbf{Performance enhancement through feedback and interaction.} We utilize \frameworkname{} to generate automated natural language feedback for previous generations, prompting the model to generate text in subsequent rounds. This approach resembles a pass@k setup but incorporates additional feedback. In Figure \ref{fig:interactive}, we observe a significant 20\% improvement in GPT-4 performance after the second round of feedback. However, the model's performance plateaus at 66\% even after three additional rounds of feedback, comparable to pass@5. The extent of performance improvement varies across tasks, with \texttt{word03}'s constraint satisfaction rate increasing from 62.1\% to 10\%. Conversely, \texttt{word02}, \texttt{sent01}, and \texttt{sent02} tasks remain challenging for the model. These findings suggest that there is still room for improvement, highlighting the difficulty of our dataset, and emphasizing the need for further research on better ways to incorporate natural language feedback.

\section{Conclusion}\label{sec:conclusion}
\vspace{-8pt}
In this work, we present \frameworkname{}, a grammar-based framework for specifying textual constraints. \frameworkname{} simplifies the process of creating constrained-generation datasets by enabling researchers to focus on specifying high level constraint structures, while \frameworkname{} automatically extracts constraint values, renders natural language instructions, and assesses model performance. 
To demonstrate the utility of the \frameworkname{} framework, we construct \datasetname{} with 1,132 constraints from 13 different types, extracted from 3 different data sources. We evaluate five state-of-the-art LLMs of various sizes on \datasetname{}, and find that it provides fine-grained insights into model capabilities and shortcomings. We hope that model developers can use \datasetname{} to develop more capable models, while future releases of \frameworkname{} can continue to adapt to the capabilities and needs of future models and users.

\textbf{Limitations and impacts.} 
Although care was taken to design the filtering and processing functions, such automated approaches are never perfect and can affect the quality of downstream constraints. We plan to continue mitigating such effects with additional utilities such as grammar checkers and parsers. Our representative constraint structures were selected to encompass diverse constrained generation challenges, but as with all generation benchmarks, they cannot capture all dimensions and nuances of model capabilities.
Benchmarks are highly influential in shaping model development, the capabilities and limitations of which may disproportionately impact different communities. Our benchmark is no exception. However, by providing an extensible, easy-to-use framework for constraint development, we hope \frameworkname{} will enable diverse stakeholders to engage with the construction of datasets, helping ensure that future model capabilities serve diverse interests and needs.
% Further data cleaning. Better class design and quantity balancing (todo). 

\section*{Acknowledgements}
We thank Xiao Liu for Vicuna/Alpaca APIs that supported our preliminary experiments, and Princeton NLP Group for helpful discussion and feedback in general. We acknowledge support from the National Science Foundation under Grant No. 2107048. Any opinions, findings, and conclusions or recommendations expressed in this material are those of the author(s) and do not necessarily reflect the views of the National Science Foundation.

% \newpage
% \clearpage

% \newpage
\bibliography{ref}

\begin{thebibliography}{38}
\providecommand{\natexlab}[1]{#1}
\providecommand{\url}[1]{\texttt{#1}}
\expandafter\ifx\csname urlstyle\endcsname\relax
  \providecommand{\doi}[1]{doi: #1}\else
  \providecommand{\doi}{doi: \begingroup \urlstyle{rm}\Url}\fi

\bibitem[Anil et~al.(2023)Anil, Dai, Firat, Johnson, Lepikhin, Passos, Shakeri,
  Taropa, Bailey, Chen, et~al.]{anil2023palm}
R.~Anil, A.~M. Dai, O.~Firat, M.~Johnson, D.~Lepikhin, A.~Passos, S.~Shakeri,
  E.~Taropa, P.~Bailey, Z.~Chen, et~al.
\newblock Palm 2 technical report.
\newblock \emph{arXiv preprint arXiv:2305.10403}, 2023.

\bibitem[Bird(2006)]{bird2006nltk}
S.~Bird.
\newblock Nltk: the natural language toolkit.
\newblock In \emph{Proceedings of the COLING/ACL 2006 Interactive Presentation
  Sessions}, pages 69--72, 2006.

\bibitem[Brooke et~al.(2015)Brooke, Hammond, and Hirst]{brooke2015gutentag}
J.~Brooke, A.~Hammond, and G.~Hirst.
\newblock Gutentag: an nlp-driven tool for digital humanities research in the
  project gutenberg corpus.
\newblock In \emph{Proceedings of the Fourth Workshop on Computational
  Linguistics for Literature}, pages 42--47, 2015.

\bibitem[Brown et~al.(2020)Brown, Mann, Ryder, Subbiah, Kaplan, Dhariwal,
  Neelakantan, Shyam, Sastry, Askell, Agarwal, Herbert-Voss, Krueger, Henighan,
  Child, Ramesh, Ziegler, Wu, Winter, Hesse, Chen, Sigler, Litwin, Gray, Chess,
  Clark, Berner, McCandlish, Radford, Sutskever, and Amodei]{brown2020gpt3}
T.~B. Brown, B.~Mann, N.~Ryder, M.~Subbiah, J.~Kaplan, P.~Dhariwal,
  A.~Neelakantan, P.~Shyam, G.~Sastry, A.~Askell, S.~Agarwal, A.~Herbert-Voss,
  G.~Krueger, T.~Henighan, R.~Child, A.~Ramesh, D.~M. Ziegler, J.~Wu,
  C.~Winter, C.~Hesse, M.~Chen, E.~Sigler, M.~Litwin, S.~Gray, B.~Chess,
  J.~Clark, C.~Berner, S.~McCandlish, A.~Radford, I.~Sutskever, and D.~Amodei.
\newblock Language models are few-shot learners.
\newblock In \emph{Advances in Neural Information Processing Systems
  (NeurIPS)}, 2020.
\newblock URL
  \url{https://papers.nips.cc/paper/2020/hash/1457c0d6bfcb4967418bfb8ac142f64a-Abstract.html}.

\bibitem[Chen et~al.(2022{\natexlab{a}})Chen, Li, Chen, and
  Narasimhan]{chen2022cognac}
H.~Chen, H.~Li, D.~Chen, and K.~Narasimhan.
\newblock Controllable text generation with language constraints.
\newblock In \emph{preprint}, 2022{\natexlab{a}}.
\newblock URL \url{https://arxiv.org/abs/2212.10466}.

\bibitem[Chen et~al.(2022{\natexlab{b}})Chen, Tworek, Jun, Yuan, Ponde~de
  Oliveira~Pinto, Kaplan, Edwards, Burda, Joseph, Brockman, Ray, Puri, Krueger,
  Petrov, Khlaaf, Sastry, Mishkin, Chan, Gray, Ryder, Pavlov, Power, Kaiser,
  Bavarian, Winter, Tillet, Petroski~Such, Cummings, Plappert, Chantzis,
  Barnes, Herbert-Voss, Hebgen~Guss, Nichol, Paino, Tezak, Tang, Babuschkin,
  Balaji, Jain, Saunders, Hesse, Carr, Leike, Achiam, Misra, Morikawa, Radford,
  Knight, Brundage, Murati, Mayer, Welinder, McGrew, Amodei, McCandlish,
  Sutskever, and Zaremba]{chen2021codex}
M.~Chen, J.~Tworek, H.~Jun, Q.~Yuan, H.~Ponde~de Oliveira~Pinto, J.~Kaplan,
  H.~Edwards, Y.~Burda, N.~Joseph, G.~Brockman, A.~Ray, R.~Puri, G.~Krueger,
  M.~Petrov, H.~Khlaaf, G.~Sastry, P.~Mishkin, B.~Chan, S.~Gray, N.~Ryder,
  M.~Pavlov, A.~Power, L.~Kaiser, M.~Bavarian, C.~Winter, P.~Tillet,
  F.~Petroski~Such, D.~Cummings, M.~Plappert, F.~Chantzis, E.~Barnes,
  A.~Herbert-Voss, W.~Hebgen~Guss, A.~Nichol, A.~Paino, N.~Tezak, J.~Tang,
  I.~Babuschkin, S.~Balaji, S.~Jain, W.~Saunders, C.~Hesse, A.~N. Carr,
  J.~Leike, J.~Achiam, V.~Misra, E.~Morikawa, A.~Radford, M.~Knight,
  M.~Brundage, M.~Murati, K.~Mayer, P.~Welinder, B.~McGrew, D.~Amodei,
  S.~McCandlish, I.~Sutskever, and W.~Zaremba.
\newblock Evaluating large language models trained on code.
\newblock In \emph{preprint}, 2022{\natexlab{b}}.
\newblock URL \url{https://arxiv.org/abs/2107.03374}.

\bibitem[Chevalier-Boisvert et~al.(2019)Chevalier-Boisvert, Bahdanau, Lahlou,
  Willems, Saharia, Nguyen, and Bengio]{chevalier-boisvert2018babyai}
M.~Chevalier-Boisvert, D.~Bahdanau, S.~Lahlou, L.~Willems, C.~Saharia, T.~H.
  Nguyen, and Y.~Bengio.
\newblock Babyai: A platform to study the sample efficiency of grounded
  language learning.
\newblock In \emph{International Conference on Learning Representations
  (ICLR)}, 2019.
\newblock URL \url{https://arxiv.org/abs/1810.08272}.

\bibitem[Chiang et~al.(2023)Chiang, Li, Lin, Sheng, Wu, Zhang, Zheng, Zhuang,
  Zhuang, Gonzalez, Stoica, and Xing]{vicuna2023}
W.-L. Chiang, Z.~Li, Z.~Lin, Y.~Sheng, Z.~Wu, H.~Zhang, L.~Zheng, S.~Zhuang,
  Y.~Zhuang, J.~E. Gonzalez, I.~Stoica, and E.~P. Xing.
\newblock Vicuna: An open-source chatbot impressing gpt-4 with 90\%* chatgpt
  quality, March 2023.
\newblock URL \url{https://lmsys.org/blog/2023-03-30-vicuna/}.

\bibitem[Dathathri et~al.(2020)Dathathri, Madotto, Lan, Hung, Frank, Molino,
  Yosinski, and Liu]{dathathri2020pplm}
S.~Dathathri, A.~Madotto, J.~Lan, J.~Hung, E.~Frank, P.~Molino, J.~Yosinski,
  and R.~Liu.
\newblock Plug and play language models: A simple approach to controlled text
  generation.
\newblock In \emph{International Conference on Learning Representations
  (ICLR)}, 2020.
\newblock URL \url{https://openreview.net/pdf?id=H1edEyBKDS}.

\bibitem[Dinu et~al.(2019)Dinu, Mathur, Federico, and
  Al-Onaizan]{dinu2019training}
G.~Dinu, P.~Mathur, M.~Federico, and Y.~Al-Onaizan.
\newblock Training neural machine translation to apply terminology constraints.
\newblock In \emph{Association for Computational Linguistics (ACL)}, 2019.

\bibitem[Foundation(2022)]{wikidump}
W.~Foundation.
\newblock Wikimedia downloads, 2022.
\newblock URL \url{https://dumps.wikimedia.org}.

\bibitem[Hamborg et~al.(2017)Hamborg, Meuschke, Breitinger, and
  Gipp]{Hamborg2017ccnews}
F.~Hamborg, N.~Meuschke, C.~Breitinger, and B.~Gipp.
\newblock news-please: A generic news crawler and extractor.
\newblock In \emph{Proceedings of the 15th International Symposium of
  Information Science}, pages 218--223, March 2017.
\newblock \doi{10.5281/zenodo.4120316}.

\bibitem[Hasler et~al.(2018)Hasler, de~Gispert, Iglesias, and
  Byrne]{hasler2018neural}
E.~Hasler, A.~de~Gispert, G.~Iglesias, and B.~Byrne.
\newblock Neural machine translation decoding with terminology constraints.
\newblock In \emph{North American Association for Computational Linguistics
  (NAACL)}, 2018.

\bibitem[Hendrycks et~al.(2020)Hendrycks, Burns, Basart, Zou, Mazeika, Song,
  and Steinhardt]{hendrycks2020measuring}
D.~Hendrycks, C.~Burns, S.~Basart, A.~Zou, M.~Mazeika, D.~Song, and
  J.~Steinhardt.
\newblock Measuring massive multitask language understanding.
\newblock \emph{arXiv preprint arXiv:2009.03300}, 2020.

\bibitem[Hokamp and Liu(2017)]{hokamp2017lexically}
C.~Hokamp and Q.~Liu.
\newblock Lexically constrained decoding for sequence generation using grid
  beam search.
\newblock In \emph{Association for Computational Linguistics (ACL)}, 2017.

\bibitem[Hu et~al.(2019)Hu, Khayrallah, Culkin, Xia, Chen, Post, and
  Van~Durme]{hu2019improved}
J.~E. Hu, H.~Khayrallah, R.~Culkin, P.~Xia, T.~Chen, M.~Post, and B.~Van~Durme.
\newblock Improved lexically constrained decoding for translation and
  monolingual rewriting.
\newblock In \emph{North American Association for Computational Linguistics
  (NAACL)}, 2019.

\bibitem[Hu et~al.(2017)Hu, Yang, Liang, Salakhutdinov, and Xing]{hu2017toward}
Z.~Hu, Z.~Yang, X.~Liang, R.~Salakhutdinov, and E.~P. Xing.
\newblock Toward controlled generation of text.
\newblock In \emph{International Conference on Machine Learning (ICML)}, 2017.

\bibitem[Johnson et~al.(2017)Johnson, Hariharan, van~der Maaten, Fei-Fei,
  Zitnick, and Girshick]{johnson2017clevr}
J.~Johnson, B.~Hariharan, L.~van~der Maaten, L.~Fei-Fei, C.~L. Zitnick, and
  R.~Girshick.
\newblock Clevr: A diagnostic dataset for compositional language and elementary
  visual reasoning.
\newblock In \emph{Conference on computer vision and pattern recognition
  (CVPR)}, 2017.
\newblock URL \url{https://ieeexplore.ieee.org/document/8099698}.

\bibitem[Keskar et~al.(2019)Keskar, McCann, Varshney, Xiong, and
  Socher]{keskar2019ctrl}
N.~S. Keskar, B.~McCann, L.~R. Varshney, C.~Xiong, and R.~Socher.
\newblock {CTRL}: A conditional transformer language model for controllable
  generation.
\newblock In \emph{preprint}, 2019.
\newblock URL \url{https://arxiv.org/abs/1909.05858}.

\bibitem[Krause et~al.(2021)Krause, Gotmare, McCann, Keskar, Joty, Socher, and
  Rajani]{krause2021gedi}
B.~Krause, A.~D. Gotmare, B.~McCann, N.~S. Keskar, S.~Joty, R.~Socher, and
  N.~F. Rajani.
\newblock {G}e{D}i: Generative discriminator guided sequence generation.
\newblock In \emph{Findings of the Empirical Methods in Natural Language
  Processing (EMNLP Findings)}, 2021.
\newblock URL \url{https://aclanthology.org/2021.findings-emnlp.424}.

\bibitem[Lhoest et~al.(2021)Lhoest, Villanova~del Moral, Jernite, Thakur, von
  Platen, Patil, Chaumond, Drame, Plu, Tunstall, Davison, {\v{S}}a{\v{s}}ko,
  Chhablani, Malik, Brandeis, Le~Scao, Sanh, Xu, Patry, McMillan-Major, Schmid,
  Gugger, Delangue, Matussi{\`e}re, Debut, Bekman, Cistac, Goehringer, Mustar,
  Lagunas, Rush, and Wolf]{huggingface-datasets}
Q.~Lhoest, A.~Villanova~del Moral, Y.~Jernite, A.~Thakur, P.~von Platen,
  S.~Patil, J.~Chaumond, M.~Drame, J.~Plu, L.~Tunstall, J.~Davison,
  M.~{\v{S}}a{\v{s}}ko, G.~Chhablani, B.~Malik, S.~Brandeis, T.~Le~Scao,
  V.~Sanh, C.~Xu, N.~Patry, A.~McMillan-Major, P.~Schmid, S.~Gugger,
  C.~Delangue, T.~Matussi{\`e}re, L.~Debut, S.~Bekman, P.~Cistac,
  T.~Goehringer, V.~Mustar, F.~Lagunas, A.~Rush, and T.~Wolf.
\newblock Datasets: A community library for natural language processing.
\newblock In \emph{Proceedings of the 2021 Conference on Empirical Methods in
  Natural Language Processing: System Demonstrations}, pages 175--184, Online
  and Punta Cana, Dominican Republic, Nov. 2021. Association for Computational
  Linguistics.
\newblock URL \url{https://aclanthology.org/2021.emnlp-demo.21}.

\bibitem[Li et~al.(2022{\natexlab{a}})Li, Yu, Khabsa, Zettlemoyer, Halevy, and
  Andreas]{li2022quantifying}
B.~Z. Li, J.~Yu, M.~Khabsa, L.~Zettlemoyer, A.~Halevy, and J.~Andreas.
\newblock Quantifying adaptability in pre-trained language models with 500
  tasks.
\newblock In \emph{North American Association for Computational Linguistics
  (NAACL)}, 2022{\natexlab{a}}.
\newblock URL \url{https://arxiv.org/abs/2112.03204}.

\bibitem[Li et~al.(2022{\natexlab{b}})Li, Thickstun, Gulrajani, Liang, and
  Hashimoto]{li2022diffusionlm}
X.~L. Li, J.~Thickstun, I.~Gulrajani, P.~Liang, and T.~B. Hashimoto.
\newblock Diffusion-lm improves controllable text generation.
\newblock In \emph{Advances in Neural Information Processing Systems
  (NeurIPS)}, 2022{\natexlab{b}}.
\newblock URL \url{https://openreview.net/forum?id=3s9IrEsjLyk}.

\bibitem[Liang et~al.(2022)Liang, Bommasani, Lee, Tsipras, Soylu, Yasunaga,
  Zhang, Narayanan, Wu, Kumar, et~al.]{liang2022holistic}
P.~Liang, R.~Bommasani, T.~Lee, D.~Tsipras, D.~Soylu, M.~Yasunaga, Y.~Zhang,
  D.~Narayanan, Y.~Wu, A.~Kumar, et~al.
\newblock Holistic evaluation of language models.
\newblock \emph{arXiv preprint arXiv:2211.09110}, 2022.

\bibitem[Lin et~al.(2020)Lin, Zhou, Shen, Zhou, Bhagavatula, Choi, and
  Ren]{lin2020commongen}
B.~Y. Lin, W.~Zhou, M.~Shen, P.~Zhou, C.~Bhagavatula, Y.~Choi, and X.~Ren.
\newblock Commongen: A constrained text generation challenge for generative
  commonsense reasoning.
\newblock In \emph{Findings of the Empirical Methods in Natural Language
  Processing (EMNLP Findings)}, 2020.

\bibitem[Lu et~al.(2021)Lu, West, Zellers, Bras, Bhagavatula, and
  Choi]{lu2021neurologic}
X.~Lu, P.~West, R.~Zellers, R.~L. Bras, C.~Bhagavatula, and Y.~Choi.
\newblock Neurologic decoding: (un)supervised neural text generation with
  predicate logic constraints.
\newblock In \emph{North American Association for Computational Linguistics
  (NAACL)}, 2021.
\newblock URL \url{https://aclanthology.org/2021.naacl-main.339.pdf}.

\bibitem[Lu et~al.(2022)Lu, Welleck, West, Jiang, Kasai, Khashabi, Bras, Qin,
  Yu, Zellers, Smith, and Choi]{lu2022neurologicAstar}
X.~Lu, S.~Welleck, P.~West, L.~Jiang, J.~Kasai, D.~Khashabi, R.~L. Bras,
  L.~Qin, Y.~Yu, R.~Zellers, N.~A. Smith, and Y.~Choi.
\newblock Neurologic a*esque decoding: Constrained text generation with
  lookahead heuristics.
\newblock In \emph{North American Association for Computational Linguistics
  (NAACL)}, 2022.
\newblock URL \url{https://aclanthology.org/2022.naacl-main.57}.

\bibitem[OpenAI(2023{\natexlab{a}})]{gpt4}
OpenAI.
\newblock Gpt-4 technical report.
\newblock In \emph{preprint}, 2023{\natexlab{a}}.
\newblock URL \url{https://arxiv.org/abs/2303.08774}.

\bibitem[OpenAI(2023{\natexlab{b}})]{openai2023gpt4}
OpenAI.
\newblock Gpt-4 technical report, 2023{\natexlab{b}}.

\bibitem[Ouyang et~al.(2022)Ouyang, Wu, Jiang, Almeida, Wainwright, Mishkin,
  Zhang, Agarwal, Slama, Ray, Schulman, Hilton, Kelton, Miller, Simens, Askell,
  Welinder, Christiano, Leike, and Lowe]{ouyang2022instructgpt}
L.~Ouyang, J.~Wu, X.~Jiang, D.~Almeida, C.~L. Wainwright, P.~Mishkin, C.~Zhang,
  S.~Agarwal, K.~Slama, A.~Ray, J.~Schulman, J.~Hilton, F.~Kelton, L.~Miller,
  M.~Simens, A.~Askell, P.~Welinder, P.~Christiano, J.~Leike, and R.~Lowe.
\newblock Training language models to follow instructions with human feedback.
\newblock In \emph{Advances in Neural Information Processing Systems
  (NeurIPS)}, 2022.
\newblock URL \url{https://openreview.net/forum?id=TG8KACxEON}.

\bibitem[Ruis et~al.(2020)Ruis, Andreas, Baroni, Bouchacourt, and
  Lake]{ruis2020gscan}
L.~Ruis, J.~Andreas, M.~Baroni, D.~Bouchacourt, and B.~M. Lake.
\newblock A benchmark for systematic generalization in grounded language
  understanding.
\newblock In \emph{Advances in Neural Information Processing Systems
  (NeurIPS)}, 2020.
\newblock URL \url{https://arxiv.org/abs/2003.05161}.

\bibitem[Schick et~al.(2023)Schick, Dwivedi-Yu, Dessì, Raileanu, Lomeli,
  Zettlemoyer, Cancedda, and Scialom]{schick2023tool}
T.~Schick, J.~Dwivedi-Yu, R.~Dessì, R.~Raileanu, M.~Lomeli, L.~Zettlemoyer,
  N.~Cancedda, and T.~Scialom.
\newblock Toolformer: Language models can teach themselves to use tools.
\newblock In \emph{preprint}, 2023.
\newblock URL \url{https://arxiv.org/abs/2302.04761}.

\bibitem[Srivastava et~al.(2022)Srivastava, Rastogi, Rao, Shoeb, Abid, Fisch,
  Brown, Santoro, Gupta, Garriga-Alonso, et~al.]{srivastava2022beyond}
A.~Srivastava, A.~Rastogi, A.~Rao, A.~A.~M. Shoeb, A.~Abid, A.~Fisch, A.~R.
  Brown, A.~Santoro, A.~Gupta, A.~Garriga-Alonso, et~al.
\newblock Beyond the imitation game: Quantifying and extrapolating the
  capabilities of language models.
\newblock \emph{arXiv preprint arXiv:2206.04615}, 2022.

\bibitem[Taori et~al.(2023)Taori, Gulrajani, Zhang, Dubois, Li, Guestrin,
  Liang, and Hashimoto]{alpaca}
R.~Taori, I.~Gulrajani, T.~Zhang, Y.~Dubois, X.~Li, C.~Guestrin, P.~Liang, and
  T.~B. Hashimoto.
\newblock Stanford alpaca: An instruction-following llama model.
\newblock \url{https://github.com/tatsu-lab/stanford_alpaca}, 2023.

\bibitem[Wang et~al.(2018)Wang, Singh, Michael, Hill, Levy, and
  Bowman]{wang2018glue}
A.~Wang, A.~Singh, J.~Michael, F.~Hill, O.~Levy, and S.~R. Bowman.
\newblock Glue: A multi-task benchmark and analysis platform for natural
  language understanding.
\newblock In \emph{Empirical Methods in Natural Language Processing (EMNLP)},
  2018.
\newblock URL \url{https://openreview.net/pdf?id=rJ4km2R5t7}.

\bibitem[Wang et~al.(2022)Wang, Mishra, Alipoormolabashi, Kordi, Mirzaei, Naik,
  Ashok, Dhanasekaran, Arunkumar, Stap, et~al.]{wang2022super}
Y.~Wang, S.~Mishra, P.~Alipoormolabashi, Y.~Kordi, A.~Mirzaei, A.~Naik,
  A.~Ashok, A.~S. Dhanasekaran, A.~Arunkumar, D.~Stap, et~al.
\newblock Super-naturalinstructions: Generalization via declarative
  instructions on 1600+ nlp tasks.
\newblock In \emph{Proceedings of the 2022 Conference on Empirical Methods in
  Natural Language Processing}, pages 5085--5109, 2022.

\bibitem[Weston et~al.(2015)Weston, Bordes, Chopra, Rush, Van~Merri{\"e}nboer,
  Joulin, and Mikolov]{weston2015towards}
J.~Weston, A.~Bordes, S.~Chopra, A.~M. Rush, B.~Van~Merri{\"e}nboer, A.~Joulin,
  and T.~Mikolov.
\newblock Towards ai-complete question answering: A set of prerequisite toy
  tasks.
\newblock \emph{arXiv preprint arXiv:1502.05698}, 2015.

\bibitem[Yao et~al.(2022)Yao, Zhao, Yu, Du, Shafran, Narasimhan, and
  Cao]{yao2022react}
S.~Yao, J.~Zhao, D.~Yu, N.~Du, I.~Shafran, K.~Narasimhan, and Y.~Cao.
\newblock React: Synergizing reasoning and acting in language models.
\newblock In \emph{International Conference on Learning Representations
  (ICLR)}, 2022.
\newblock URL \url{https://arxiv.org/abs/2210.03629}.

\end{thebibliography}
\bibliographystyle{abbrvnat}
\newpage
\clearpage
% \input{checklist}
% \newpage
% \clearpage
% \beginsupplemental
\appendix
% \section*{Appendix}
% supplementary figs

\section*{Code Availability}

The code, data, and additional details are available at\\ \url{https://collie-benchmark.github.io}.

\section{Natural Language Rendering of Constraints}\label{sec:sup:rendering}

\subsection{Instruction rendering} \label{appendix:render}
\frameworkname{} provides a rule-based constraint renderer that converts constraints into natural language instructions (see examples in Table \ref{tab:example_constraints}).

Leveraging the compositionality of the context-free grammar, the renderer first parses the constraint as a tree. In the case of multi-constraints, it generates prompts for each base-constraint individually and then concatenates the generated prompts together at the end. For a base-constraint, it follows a pre-order traversal of the subtree to modify the initial template "Please generate a \texttt{\{generation\_level\}} with @... \texttt{\{tagert\_level\}}," where "@..." serves as a placeholder.

Although the rule-based instruction prompts are natural enough for all examples in \frameworkname-v1, there might be some edge cases where the rule-based instructions are not fluent enough for newly specified constraints. To address this, we offer an option to utilize language models to enhance the rule-based instructions. We employ the following prompt for the language model to refine the instructions: "Please rewrite the following paragraph to improve fluency without altering the original meaning. You should provide the revised paragraph directly. Original paragraph: \texttt{\{prompt\}}."

The renderer is independent of the constraint construction and can be easily extended with new rules for parsing and mapping to instruction templates.

\subsection{Feedback rendering} \label{appendix:feedback}

We further extend our framework to provide natural language feedback when the extracted value of the generated text differs from the target value. Similar to instruction rendering, we first employ a rule-based renderer to compose the feedback by modifying a template. We also provide an option to use language models to polish the generated feedback.

For instance, consider a constraint $\texttt{count}(\xi, \text{char}, `v') = 2 \wedge \texttt{count}(\xi, \text{char}, `i') = 3$, while the generated word includes three `v' and four `i'. Our framework can generate the following instructions and feedback:

\begin{itemize}
	\item \texttt{INSTRUCTION}:\\
Please generate a word with exactly 2 character `v' and exactly 3 character `i'.
	\item \texttt{GPT-POLISHED INSTRUCTION}:\\
Please generate a word that contains exactly 2 instances of the letter `v' and exactly 3 instances of the letter `i'.
	\item \texttt{FEEDBACK}:\\
Your task is to generate a word with exactly 2 character `v' and exactly 3 character `i'.
However, you generate a word with 3 character `v' and 4 character `i'.
	\item \texttt{GPT-POLISHED FEEDBACK}:\\
Your task was to generate a word with precisely 2 `v' characters and precisely 3 `i' characters. However, you generated a word with precisely 3 `v' 
characters and precisely 4 `i' characters.
\end{itemize}

By incorporating this feedback mechanism, our framework can provide explicit guidance for the language models to improve the generation quality and adhere to the specified constraints.
\section{Dataset} \label{appendix:datasets}
\subsection{Extraction overview} \label{appendix:extract}
The extraction phase is split into six steps:
\begin{enumerate}
    \item \textbf{Document loading.} The document $d$, usually consisting of multiple paragraphs is loaded.
    \item \textbf{Text chunking.} Each document is divided into paragraphs using a source-specific delimiter (e.g. \textbackslash n). Sentence-level constraints are further sentence tokenized using {\tt nltk}. 
    \item \textbf{Text filtering.} The paragraph or sentence passes a source-specific filtering function that attempts to remove all strings that are not natural language, for instance copyright statements.
    \item \textbf{Text post-processing.} The paragraphs or sentences that survive filtering are post-processed to remove source-specific artifacts, such as Markdown formatting. 
    \item \textbf{Passage construction (passage-level only).} Paragraphs and sentences pass through to the next step. Passages are constructed by appending as many consecutive paragraphs that survive filtering as possible. For instance, if a document contains paragraphs $p_1,...,p_9$, and $p_4$ is the only paragraph that is removed due to filtering, then we return two passages: $(p_1,p_2,p_3)$ and $(p_5,...,p_9)$. Each paragraph is joined by two newline characters within each passage.
    \item \textbf{Constraint extraction.} The sentence, paragraph, or passage-level string is passed to the constraint extractor that pulls out constraint targets from the string. This can either be done directly, such as directly extracting the total word count, or sweeping over a set of possible target values. 
\end{enumerate}
For each data source, we randomly sample $300$ documents. For each document, we randomly sample up to $100$ text sequences of the specified level (sentence, paragraph, or passage) for constraint extraction to prevent over-representation from very long documents. We then randomly sample up to $100$ constraint targets for each constraint structure and data source. We now discuss source-specific details below:

\subsection{Text filters} \label{appendix:filter}
In this section, we describe the text filter heuristics in detail. Note that which filters to use are source-specific.
\begin{itemize}
    \item \textbf{URL.} This filter removes any string that contains a pattern that appears to be a URL. The pattern we find is expressed using the following regex:
    
    {\tt r"(http(s)?://)?(www\textbackslash{}.)?[a-zA-Z0-9\textbackslash{}-]+\textbackslash{}.[a-zA-Z]\{2,6\}}
    
    {\tt (\textbackslash{}.[a-zA-Z]\{2,6\})?(/[a-zA-Z0-9\textbackslash{}-]*)*(\textbackslash{}?[a-zA-Z0-9\textbackslash{}-=\&]*)?"}
    \item \textbf{All caps.} This filter removes any text that only contains capitalized letters, which may be indicative of a section heading.
    \item \textbf{No sentences.} This is a filter that tries to detect strings without any valid sentences in the text. We first sentence tokenize the string. If no ``sentence'' contains a period and has length greater than 2, then we remove the string. Otherwise, we keep it. Future improvements could use a parser or trained classifier.
    \item \textbf{Copyright.} This filter removes copyright statements typically found at the end of articles. It removes any string that contains the copyright symbol ``©'' or where the uncased first word is ``copyright''.
    \item \textbf{Caption.} This filter attempts to remove captions, such as those under diagrams or images. These strings typically follow the format: ``Photo: a green car.''. We heuristically detect such strings by rejecting any string where the number of characters to the left of the first ``:'' is less than six characters.
\end{itemize}

\subsection{Text post-processing} \label{appendix:postprocess}
In this section, we describe the post-processing functions used on the strings. Note that which post-processing functions to use depends on the data-source.
\begin{itemize}
    \item \textbf{Markdown removal.} We remove markdown artifacts using the following substitution rule: 
    {\tt (r'(\textbackslash{}*\textbackslash{}*|\_\_|\textbackslash{}*|\_|\text
    backslash{}\textasciitilde{}\textbackslash{}\textasciitilde{})(.*?)\textbackslash{}1', r'\textbackslash{}2')}

    \item \textbf{Consecutive whitespace.} Consecutive whitespace is removed with the following substitution rule:
    {\tt (r'\textbackslash{}s\{2,\}', ' ')}

    \item \textbf{Single newline to space.} Single newlines are replaced with a single space using the following substitution rule: {\tt (r"(?<!\textbackslash{}n)\textbackslash{}n(?!\textbackslash{}n)", " ")}

    \item \textbf{Bracket removal.} We remove brackets from the text using the following substitution rule: {\tt (r'\textbackslash{}[[\^{}\textbackslash{}]]*\textbackslash{}]', "")}. This is useful for removing references inside the text.
\end{itemize}

\subsection{Data sources} \label{appendix:sources}
Detailed statistics on the number of constraints extracted for each constraint structure for the grouped, and individual data sources are found in Figures \ref{sup-fig:grouped-sample-size} and \ref{sup-fig:sample-size} respectively. None of the datasets used contain PII, as far as authors are aware.

\paragraph{Wiki} We use the {\tt 20220301.en} train split of the dataset from Huggingface \cite{huggingface-datasets}. We split each document into paragraphs using two newlines as the delimiter ({\tt \textbackslash n\textbackslash n}). We use three filters for Wiki: URL, caption, and no sentences. For our passage level constraint, we also omit any text that contains the vertical line character "|", as these were identified to often be tables. We use a Wiki specific post-processing function that removes any text before the first newline character, for any text that contains a newline character. We found that these are almost always section headings. Wiki is licensed under a CC BY-SA 3.0 license and GNU Free Documentation License\footnote{See \url{https://dumps.wikimedia.org/legal.html}}.

\paragraph{CC-News} We load from the train split of the {\tt cc\_news} dataset on Huggingface Datasets for convenience. We split each document into paragraphs using a single newline as the delimiter ({\tt \textbackslash{}n}. We use four filters for CC-News: copyright, URL, cpation, and no sentences. We do not use any post-processing function. The TOS for this data can be found at \url{https://commoncrawl.org/terms-of-use/full/}.

\paragraph{Guten} We use the processed dataset from \textit{Gutenberg, dammit}: \url{https://github.com/aparrish/gutenberg-dammit}. We split each document using two newlines as the delimiter ({\tt \textbackslash n\textbackslash n}). We two filters for Guten: all caps, and no sentences. We post-process the text using four processors, applied in the following order: markdown removal, bracket removal, single newline to space, consecutive whitespace to single whitespace. All documents in Guten are in the public domain in the U.S.

\paragraph{Words} For word-level constraints, we iterate over the the words present in the following newline-separated word list: \url{ http://www.gwicks.net/textlists/english3.zip}. We conduct no filtering or post-processing on the words from the list.

Our entire code, including those used for data extraction will be released under an MIT license.
% \section{Limitations and Impacts}

% Although care was taken to design the filtering and processing functions, such automated approaches are never perfect and can affect the quality of downstream constraints. We plan to continue mitigating such impacts by adding additional filtering and processing utilities such as grammar checkers and parsers. Our representative constraint structures were selected to encompass many diverse constrained generation challenges, but as with all generation benchmarks, they cannot capture all dimensions and nuances of model capabilities.

% Benchmarks are highly influential in shaping model development, the capabilities and limitations of which may disproportionately impact different communities. Our benchmark is no exception. However, by providing an extensible, easy-to-use framework for constraint development, we hope that \frameworkname{} will enable diverse stakeholders to engage with the construction of datasets, helping to ensure that future model capabilities serve diverse interests and needs.
\section{Additional Experimental Results}\label{sec:sup:results}

We note that out of $1132$ constraints, $2$ constraint prompts are blocked by PaLM-2 API for the guardrailing reason: 
\begin{enumerate}
    \item In \texttt{ccnews\_c07}: Please generate a sentence containing the word 'charged', 'been', 'Father'.
    \item In \texttt{ccnews\_c14}: Please generate a passage with all paragraphs having the last sentence to be 'Gramercy's portfolio looks attractive relative to peers', 'I am going to add Gramercy Property Trust to my income portfolio this week', 'An investment in GPT yields $6.6$ percent', 'The REIT's shares have slumped a whopping $\sim 15$ percent in 2018, but are no longer oversold' respectively. Only generate the passage, without extra things like ``Paragraph 1" or ``Answer:".
\end{enumerate}

\subsection{One-shot experiments} \label{appendix:oneshot}

To understand if LLM performances are bottlenecked by the zero-shot instruction format and if example input-output pairs could boost performances, we did a preliminary one-shot prompting experiment using an internal version of the dataset before the current \datasetname{}, where for each constraint structure, we use a fixed constraint and its corpus example as an example input-output pair attached before the constraint prompt. As shown below in Table~\ref{tab:oneshot}, results are very similar for both GPT-3.5 and GPT-4, which suggests the task difficulty is mainly about generation under the constraint instead of understanding the constraint (by similar examples).

\begin{table}[ht]
    \centering
    \begin{tabular}{c|cc}
        \toprule 
         & GPT-3.5 & GPT-4 \\
         \midrule
        0-shot & 23.1 & 40.7 \\
        1-shot & 23.6 & 39.4 \\
        \bottomrule
    \end{tabular}
    \caption{0 vs.\,1-shot results across all constraints (\%).}
    \label{tab:oneshot}
\end{table}

\subsection{Constraint satisfaction rates}

\begin{figure}[ht]
    \centering
    \includegraphics[width=\textwidth]{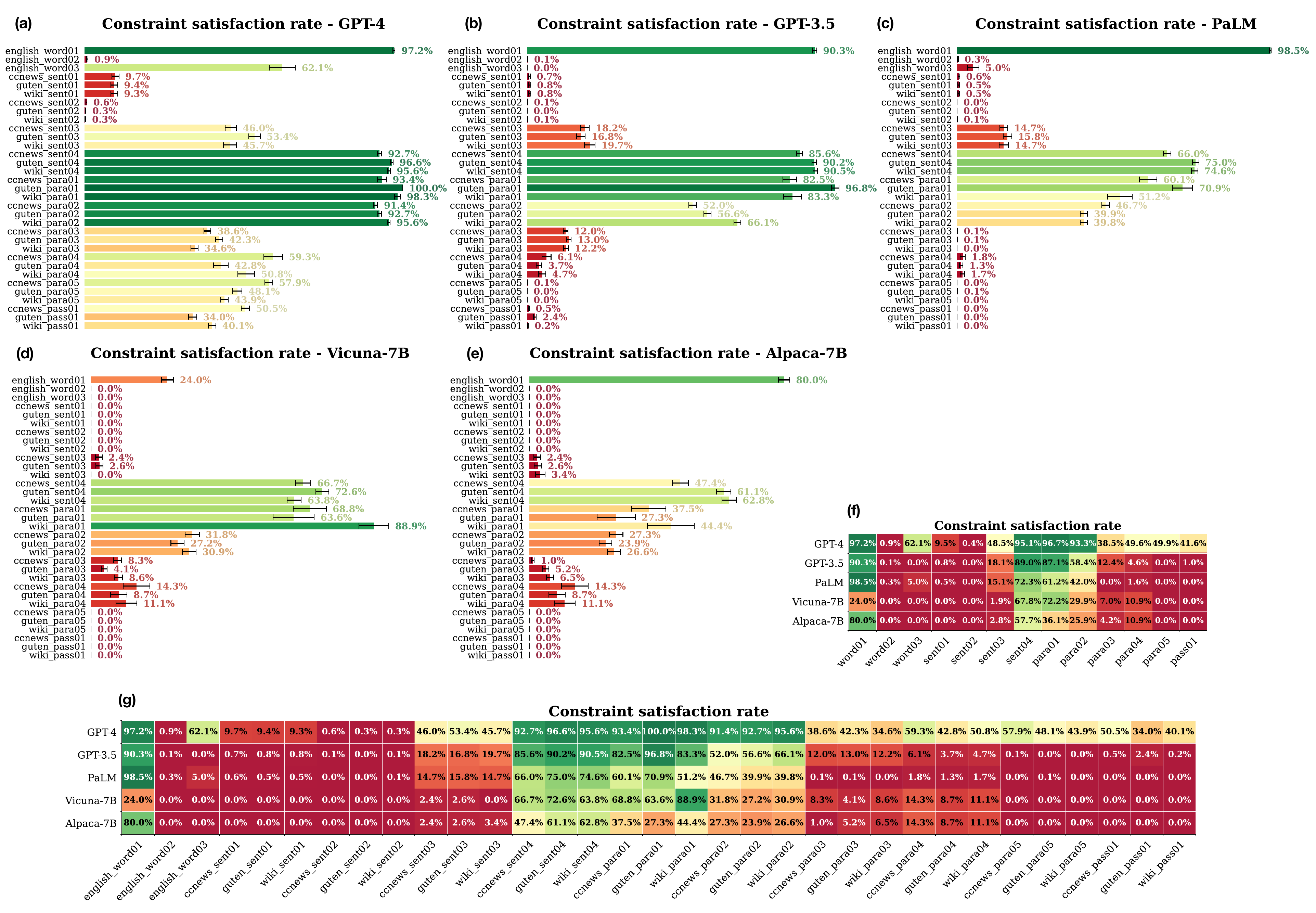}
    \caption{\textbf{Model performance on different constraints and datasets.} \textbf{(a)-(e)} Constraint satisfaction rates of texts generated by GPT-4, GPT-3.5, PaLM, Vicuna-7B, and Alpaca-7B across various constraints and datasets. Error bars indicate standard error. The constraint group names can be found in Table \ref{tab:example_constraints}. Sample sizes are reported in supplementary Figure \ref{sup-fig:sample-size}. \textbf{(f)} Summary heatmap of model performance on different constraint groups. \textbf{(g)} Summary heatmap of model performance on different constraints and datasets.}
    \label{sup-fig:performance-comparison}
\end{figure}

Figure \ref{sup-fig:performance-comparison} provides a comparison of constraint satisfaction rates for various models across all tasks. The performance of the models remains consistently high for a specific constraint structure, regardless of the data source. The satisfaction rates are summarized in heatmap Figures \ref{sup-fig:performance-comparison}(f)-(g).

\begin{figure}[ht]
    \centering
    \includegraphics[width=\textwidth]{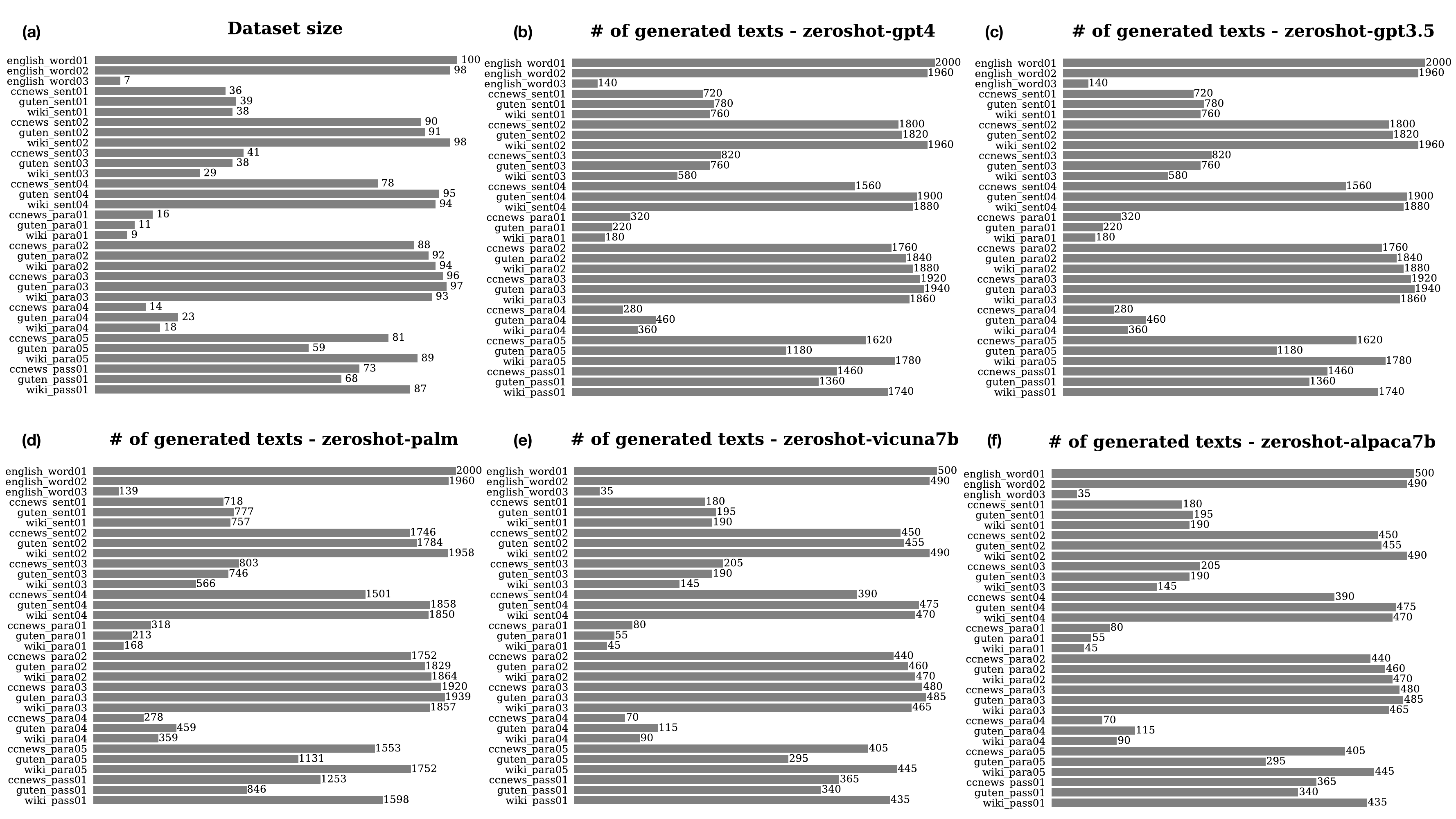}
    \caption{\textbf{Dataset and sample sizes.} \textbf{(a)} Dataset sizes for each constraint and data source. \textbf{(b)-(f)} Total sample sizes of generated texts from different models for each constraint and data source.}
    \label{sup-fig:sample-size}
\end{figure}

\begin{figure}[ht]
    \centering
    \includegraphics[width=\textwidth]{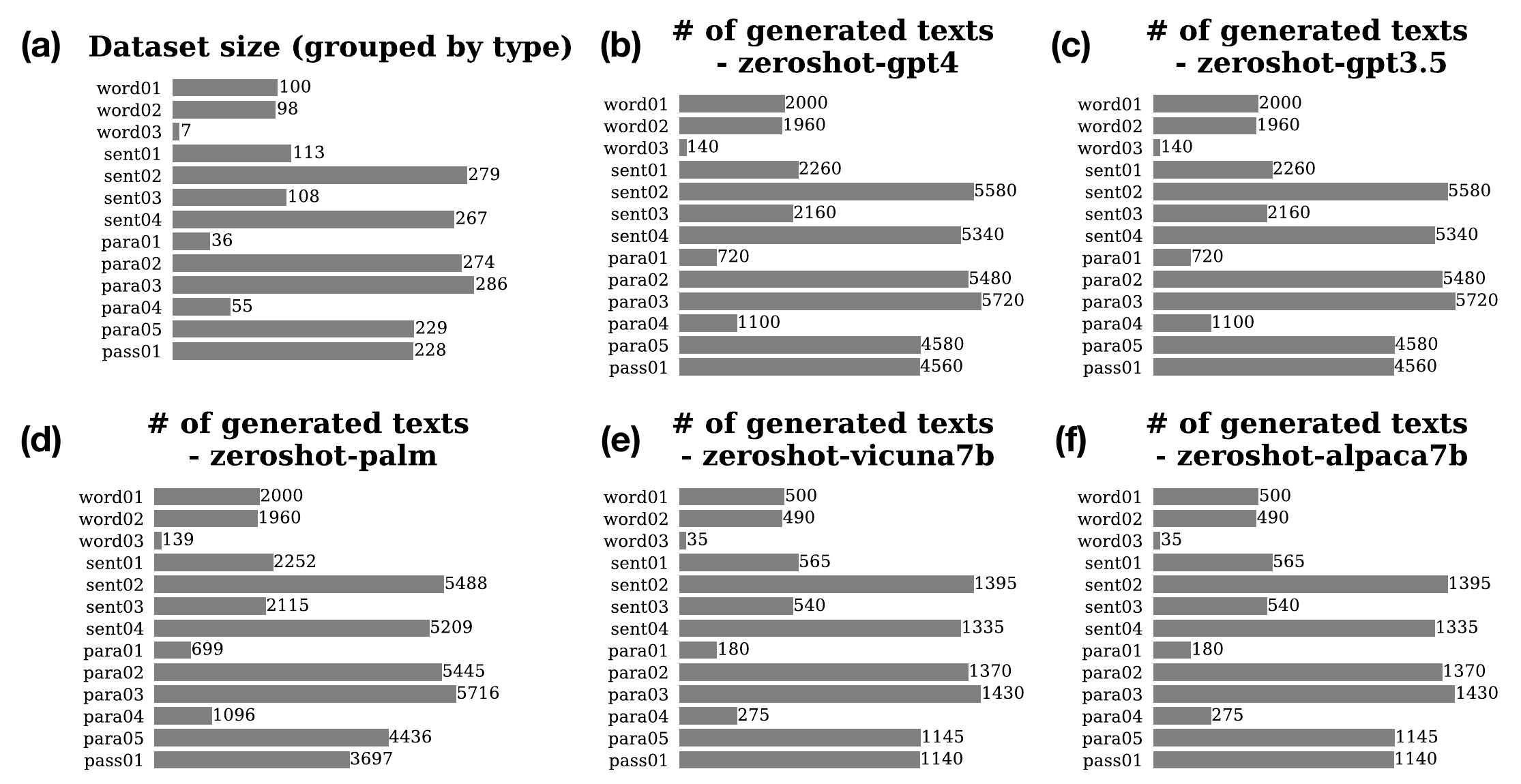}
    \caption{\textbf{Grouped dataset and sample sizes.} \textbf{(a)} Dataset sizes for each constraint group. \textbf{(b)-(f)} Total sample sizes of generated texts from different models for each constraint group.}
    \label{sup-fig:grouped-sample-size}
\end{figure}

Figures \ref{sup-fig:sample-size} and \ref{sup-fig:grouped-sample-size} provide detailed information about the dataset size and sample size for each model in the study. Specifically, we conducted 20 trials for each instruction prompt in the case of GPT-4 and GPT-3.5. For PaLM, a total of 30 trials were conducted for each instruction prompt. However, due to a certain failure rate, the number of generated texts may not be a multiple of the number of instruction prompts. Vicuna-7B and Alpaca-7B were each run for 10 trials, and they also experienced some low failure rates during the experiments.

\subsection{Additional evaluations}

In addition to evaluating binary constraint satisfaction in general, it is also possible to evaluate particular aspects of text generation with respect to the constraint and extracted corpus text.

\textbf{Word validity evaluation.} To determine the validity of word-level generations as English words, we cross-reference the generated words with the word list available at \url{http://www.gwicks.net/textlists/english3.zip}. Since this word list is not complete, we supplement it by including eight additional uncommon but valid English words: 'supercalifragilisticexpialidocious', 'pneumonoultramicroscopicsilicovolcanoconiosis', 'antidisestablishmentarianism', 'pseudopseudohypoparathyroidism', 'extraterrestrializationism',
'acceleratrix', 'circumlocutrix', and 'procrastinatrix'.

\begin{figure}[ht]
    \begin{floatrow}
        \ffigbox[\FBwidth]
            {\caption{\textbf{Word validity.} Percentage of generated words that are ``valid" words for a given English vocabulary list. Language models can sometimes generate plausible words, such as ``coordinasor" and ``adventudposis", but those are not common or valid words in modern English.}
    \label{sup-fig:word-validity}}
            {\includegraphics[width=0.48\textwidth]{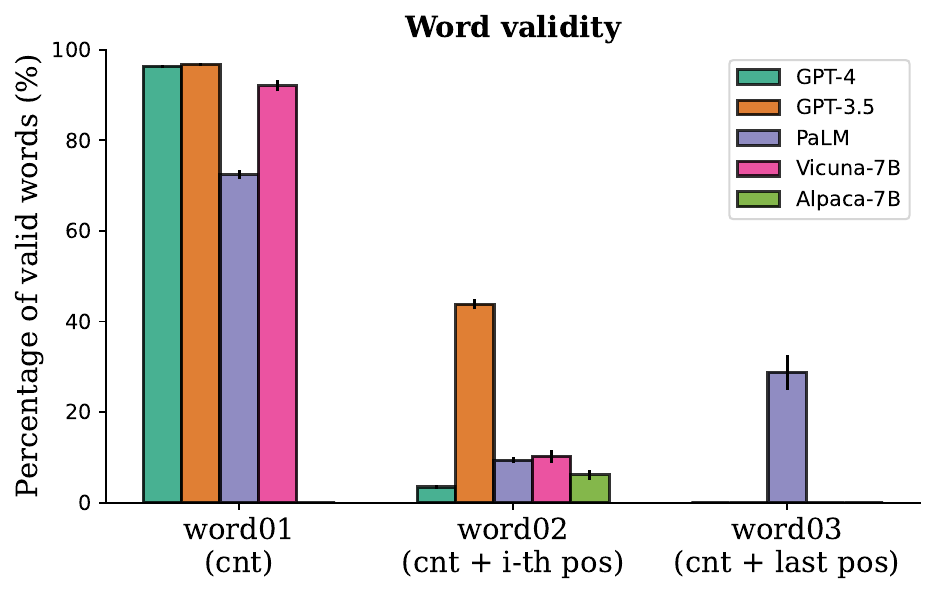}}
        \ffigbox[\FBwidth]
            {\caption{\textbf{Passage coherence.} Average passage coherence scores rated by GPT-4. Each generated passage was evaluated through three independent runs, while roughly one trial was taken for each model. GT-examples are ground truth. The error bars represent the standard error across the dataset.}\label{sub-fig:pass-coherence}}
            {\includegraphics[width=0.48\textwidth]{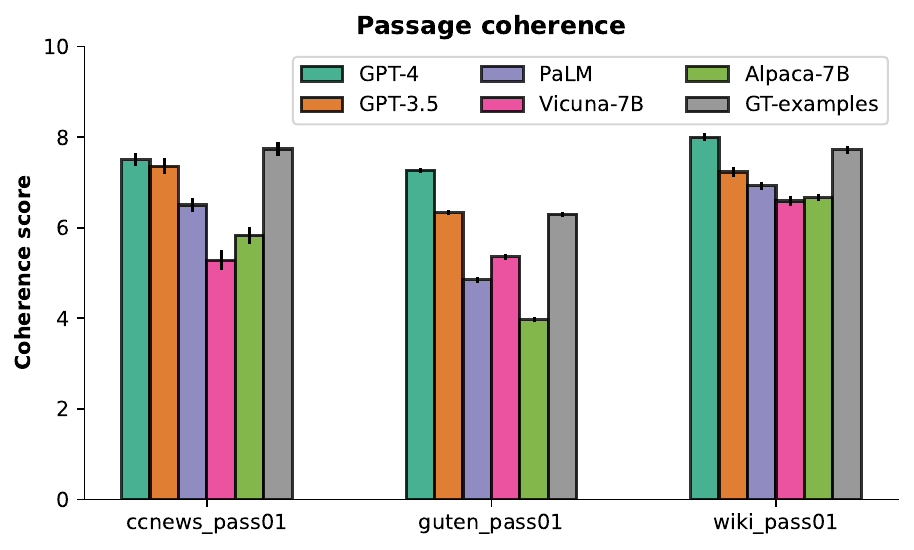}}
    \end{floatrow}
    % \vspace{-1em}
\end{figure}

Figure \ref{sup-fig:word-validity} illustrates the performance comparison of different models in generating long words (\texttt{word01}). Notably, GPT-4 demonstrates superior performance compared to other models in generating long words. However, when faced with more challenging constraints, such as the requirement for the $i$-th letter to be 'r', all models fail to generate a word that satisfies the constraint (see Figure \ref{sup-fig:performance-comparison}). In this case, GPT-3.5 manages to generate valid words, while GPT-4 resorts to fabricating words like "coordinasor" to better conform to the constraints.

Regarding task \texttt{word03}, only GPT-4 is capable of generating words that satisfy the constraint on the last character. However, it still frequently generates made-up words. None of the other models are able to generate valid words or strings that satisfy the given constraint.

\textbf{Passage coherence evaluation.} In order to assess the coherence and flow of content within the generated paragraphs, we utilize GPT-4 as a third-party judge to provide coherence scores. For this evaluation, we employ the following prompt: "Analyze the following passage, then conclude with the statement 'Thus, the coherency score is {$s$},' where $s$ is an integer ranging from 1 to 10." We conduct three separate samplings of coherence scores for each generated text and calculate the average score. This methodology allows us to quantitatively measure the overall coherence of the generated paragraphs and gauge their coherence in a relatively consistent and reliable manner.

Figure \ref{sub-fig:pass-coherence} presents the coherence scores of generated passages for task \texttt{pass01}. Notably, both GPT-4 and GPT-3.5 consistently outperform the other models in terms of coherence. Furthermore, GPT-4 achieves a level of coherence that is comparable to the ground truth passages in the dataset.

\end{document}